\documentclass[acmsmall]{acmart}

\AtBeginDocument{%
  \providecommand\BibTeX{{%
    \normalfont B\kern-0.5em{\scshape i\kern-0.25em b}\kern-0.8em\TeX}}}

\setcopyright{acmlicensed}
\acmJournal{TKDD}
\acmYear{2021} \acmVolume{1} \acmNumber{1} \acmArticle{1} \acmMonth{1} \acmPrice{15.00}\acmDOI{10.xxx/xxx}

\acmJournal{TKDD}
\acmVolume{1}
\acmNumber{1}
\acmArticle{111}
\acmMonth{1}

\usepackage{graphicx}
\usepackage{subfigure}
\usepackage{bm}
\usepackage{amsmath}
\usepackage{algorithm}
\usepackage{algorithmic}
\usepackage{array}

\usepackage{tabularx}
\usepackage{color}
\usepackage{subfigure}

\begin{document}

\title{Learning Graph Neural Networks with Positive and Unlabeled Nodes}

\author{Man Wu}
\affiliation{%
  \institution{Florida Atlantic University, USA}
}
\email{mwu2019@fau.edu}

\author{Shirui Pan}
\affiliation{%
  \institution{Monash University, Australia}
}
\email{shirui.pan@monash.edu}

\author{Lan Du}
\affiliation{%
  \institution{Monash University, Australia}
}
\email{lan.du@monash.edu}


\author{Xingquan Zhu}
\affiliation{%
  \institution{Florida Atlantic University, USA}
}
\email{xzhu3@fau.edu}

\renewcommand{\shortauthors}{Wu et al.}

\begin{abstract}

Graph neural networks (GNNs) are important tools for transductive learning tasks, such as node classification in graphs, due to their expressive power in capturing complex interdependency between nodes. To enable graph neural network learning, existing works typically assume that labeled nodes, from two or multiple classes, are provided, so that a discriminative classifier can be learned from the labeled data. In reality, this assumption might be too restrictive for applications, as users may only provide labels of interest in a single class for a small number of nodes. In addition, most GNN models only aggregate information from short distances (\textit{e.g.}, 1-hop neighbors) in each round, and fail to capture \textit{long distance relationship} in graphs. In this paper, we propose a novel graph neural network framework, long-short distance aggregation networks (\texttt{LSDAN}), to overcome these limitations. By generating multiple graphs at different distance levels, based on the adjacency matrix, we develop a long-short distance attention model to model these graphs. The direct neighbors are captured via a short-distance attention mechanism, and neighbors with long distance are captured by a long distance attention mechanism. Two novel risk estimators are further employed to aggregate long- short-distance networks, for PU learning and the loss is back-propagated for model learning. Experimental results on real-world datasets demonstrate the effectiveness of our algorithm.
  %
  %
\end{abstract}



\maketitle


\section{Introduction}
With the rapid development of networking platforms and data intensive applications, graphs are becoming convenient and fundamental tools to model the complex inter-dependence among big scale data. As a result, networks (or graphs) are being widely used in many applications, including citation networks~\citep{Kipf2016Semi}, social media networks~\citep{aclBaldwinCR18}, webpage networks~\citep{Chen2017Graph}, protein-protein interaction networks~\citep{NIPS2017_7231} and so forth. 
Graph data, however, is inherently sparse and highly complex, making it difficult to carry out graph analytic tasks.
For example, graph node classification attempts to categorize nodes in a network into a number of groups, where the essential challenge is the integration of both the graph structure and the node content information. 

In order to capture node content and graph structure, many approaches have been proposed recently to embed both structure and node content information of graphs into a compact and low dimensional space for a new representation learning.
These existing methods can be roughly categorized into two groups: (1) two-step graph embedding based classification algorithms, and (2) end-to-end graph convolutional neural nets methods.

For two-step graph embedding methods, graph embedding based algorithms first embed nodes in a given graph into vector representation by preserving both structure, node content, and other side information. Then a classical supervised learning algorithm, such as support vector machine, is built from the vector data for classification. Graph embedding algorithms are often learned in an unsupervised manner. They  either capture the walk-based similarity between nodes, such as DeepWalk~\citep{perozzi2014deepwalk}, LINE~\citep{tang2015line}, node2vec~\citep{kdd_GroverL16}, or apply autoencoder-based models to reconstruct the graph structure information, such as DNGR~\citep{cao2016deep} and SDNE~\citep{Wang_SDNKDD2016}. 
While being relatively simple, one limitation of these models is that they separate the embedding and the classification task into two steps. As a result, the learned node features may not have best representation for the succeeding classifiers to learn an effective discriminitive model for the node classification task.

\begin{figure}[t]
\centering\includegraphics[width=0.48\textwidth]{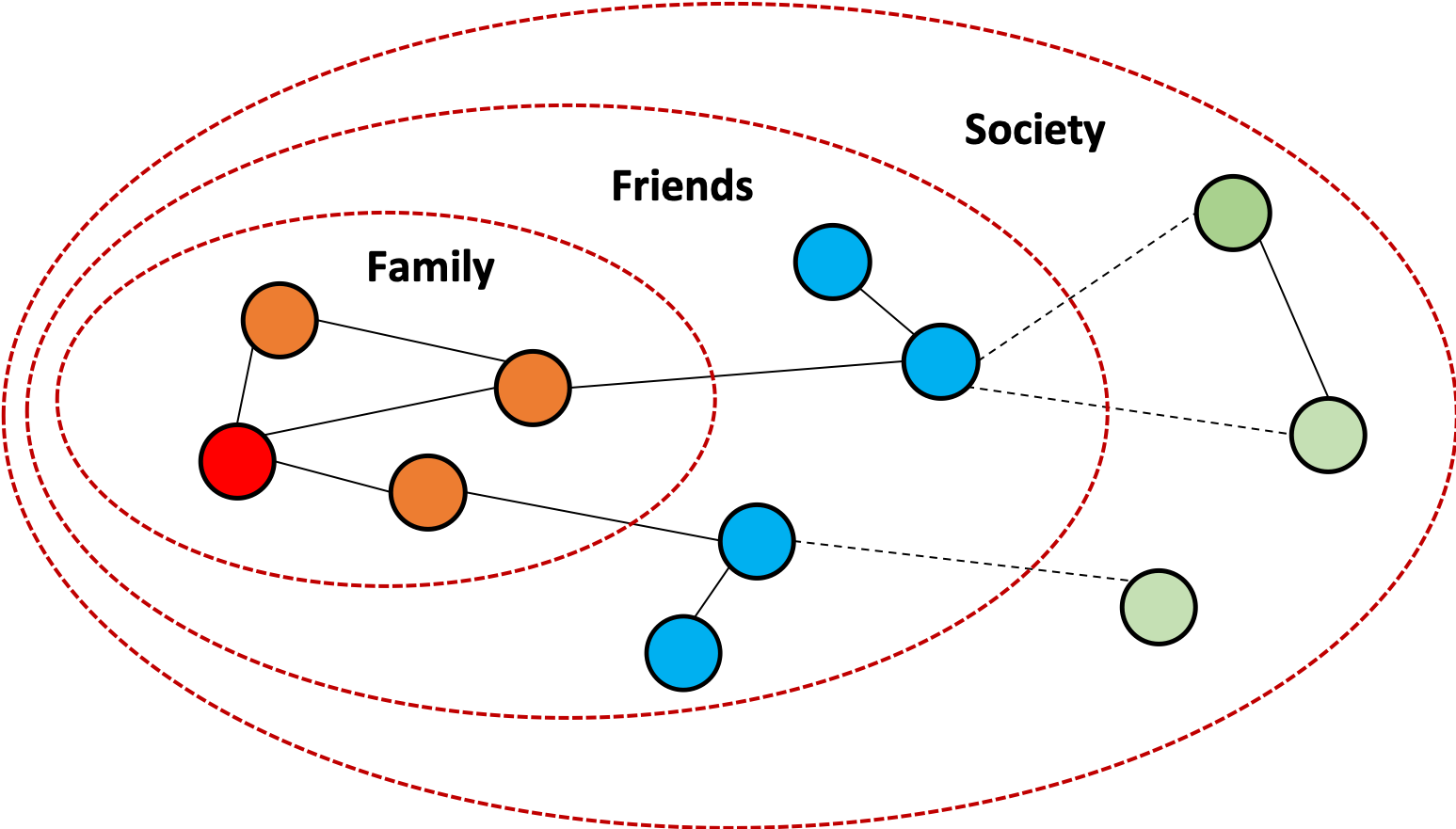}
\caption{A conceptual view of short and long distance relationship. Family members are considered short distance relationships and society members are considered long distance relationships.}
 \label{fig:intro}
\end{figure}

On the other hand, graph neural network approaches, such as graph convolutional networks (GCNs) \citep{Kipf2016Semi}, employ an end-to-end framework to overcome the limitation of two-step approaches and have achieved impressive performances in the node classification task. The essential idea of GCNs is to generate a \textit{convolutional layer} to exploit the irregular graph structure information and utilize a classification loss function to assist an attributed graph accomplish the classification task.
The \textit{graph convolution} operation is described as a filtering process aggregating features from neighboring nodes, \textit{i.e.},
\begin{equation}
    \setlength{\abovedisplayskip}{1mm}
    \setlength{\belowdisplayskip}{1mm}
   h_i^{(\mathrm{l}+1)} = h_i^{(\mathrm{l})} + \texttt{Aggregate} (h_j)_{j \in \Gamma_{i}},
   \label{eq:ave_agg}
\end{equation}
here, $h_i^{(\mathrm{l}+1)}$ and $h_i^{(\mathrm{l})}$ are latent feature representations of the node $v_i$ at the $\mathrm{l}+1$ and $\mathrm{l}$-th layer, respectively, and  \texttt{Aggregate($\cdot$)} is an operation that assorts information from neighbor nodes ($\Gamma_{i}$) of vertex $v_i$. 

After obtaining the new information, GCN applies a neural network to learn a new representation  via $o_i^{(\mathrm{l}+1)}= W  h_i^{(\mathrm{l}+1)}$ through a learnable weight matrix $W$. More specifically, GCN and GraphSage \citep{hamilton2017inductive} define the \texttt{Aggregate($\cdot$)} as the average, $i.e., \texttt{Ave} (h_j)_{j \in \Gamma_(i)}$, or summarization of neighboring feature information that equally considers the significance of each neighbor in the learning process. The recent proposed graph attention network (GAT) aims to learn the weights of different neighbors for aggregating information \citep{velickovic2017graph}.
%
\subsection{Motivation}

Although commonly used, one of the key limitations of graph neural networks (including GAT \citep{velickovic2017graph}) is that they rely on direct (1-hop) neighbor nodes to learn weight information. As a result, \textit{long distance relationship} is largely ignored in the representation learning process~\cite{wu2019cikm}. 

In practice, \textit{long distance relationship} is vitally significant. 
%
For example, in social networks, an individual is influenced by her/his neighborhood relations at different distance levels, ranging from short distance relationships (\textit{e.g.} families, friends), to long distance relationships (\textit{e.g.} society, nation states). 
An example of long distance relationship is illustrated in Figure \ref{fig:intro}. 
Since every single relationship is generally sparse and biased, long distance relationship should be additionally considered for graph learning to obtain a comprehensive representation of each node collaboratively.

\begin{figure}[t]
\centering\includegraphics[width=0.8\textwidth]{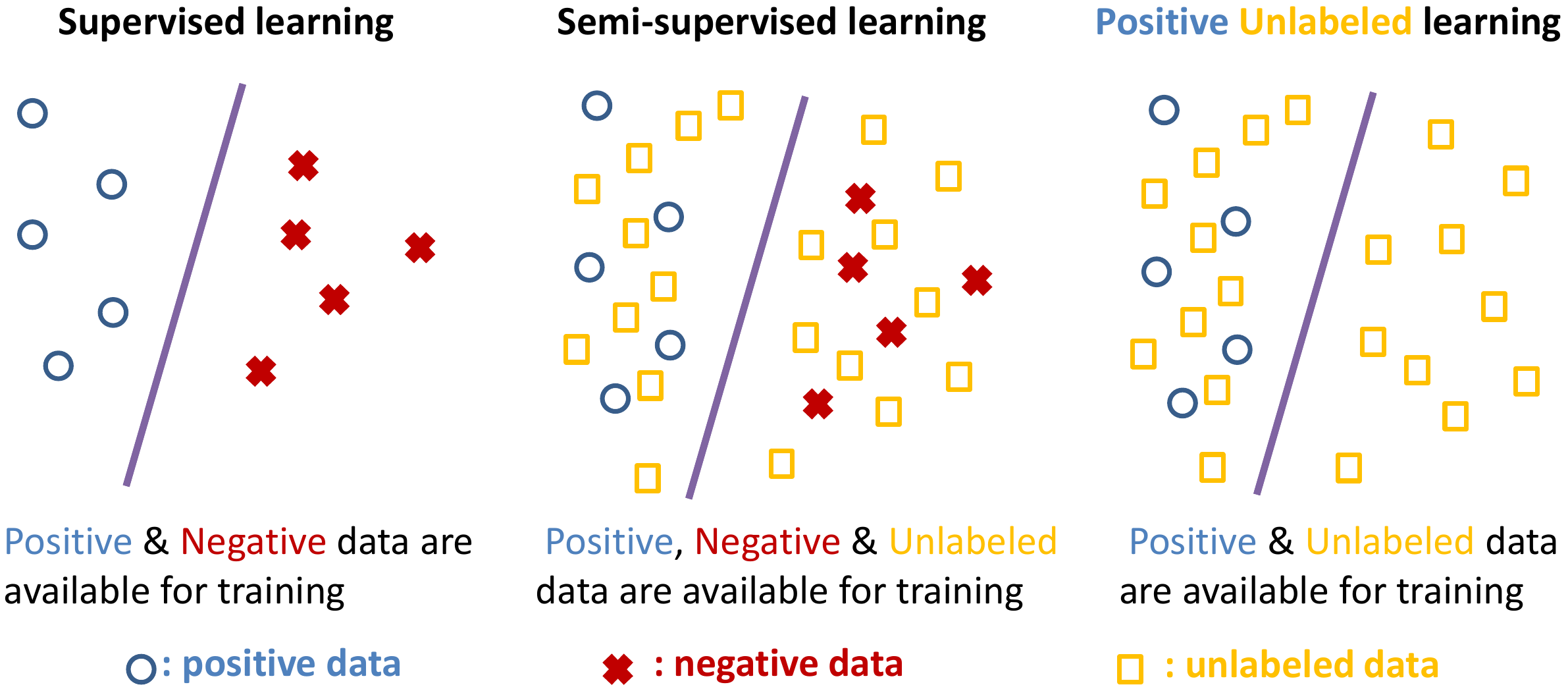}
\caption{An example of supervised learning, semi-supervised learning, and positive unlabeled learning. Supervised learning and semi-supervised have two or more types (classes) of labeled instances, whereas positive unlabeled learning only has one type of labeled instances (\textit{i.e.} positive samples).}
 \label{fig:intro2}
\end{figure}

In addition to the long-short distance relationships, another limitation of existing graph neural networks is that they require users to label data from two or more classes to help facilitate the classification task.  
This is because that most existing graph neural networks, such as GCN and GAT, are supervised learning or semi-supervised learning approaches, where training samples should include labeled positive and negative samples for binary classification tasks or more types of labeled samples for multi-class classification tasks. Such requirement inevitably imposes significant labeling costs, and in some cases, 
%
users may only provide labels of interest for a small number of nodes in one specific class. 
%
%
For example, when surfing the Internet, which is an enormous graph, users may only bookmark pages interesting to them and ignore rest of pages. As a result, only positive samples (\textit{i.e.} bookmarked pages) are labeled and all other pages are unlabeled.  

%
The above observations show a positive unlabeled learning problem setting to recommend pages or news of interest to users. An example of the positive unlabeled learning, compared to supervised learning and semi-supervised learning, is illustrated in Figure~\ref{fig:intro2}. As graphs are becoming increasingly popular in applications, many methods are replying on graph neural networks and graph attention mechanisms for learning and analysis. 
Although positive and unlabeled learning have been previously studied for generic data~\cite{Li2003Learning} and graphs~\cite{wu2017positive}, this problem has not been addressed and explored by existing graph neural networks.

Motivated by the above observation, our research intends to leverage long-short distance relationships and design new graph neural network approaches for positive and unlabeled graph learning.

\subsection{Challenges and Contribution}
In this paper, we explore the positive unlabeled graph neural network learning, in which only partial positive nodes are labeled.
%
Considering the extensive usage of graph neural networks as learning frameworks in previous study\citep{Kipf2016Semi,hamilton2017inductive}, we summarize following two main challenges:

\begin{itemize}
\item \textit{Challenge 1}: How to capture graph structure information from long-distance neighbors? Typically, existing graph neural networks only utilize short-distance information in a single layer.
\item \textit{Challenge 2}: How to design an end to end framework for positive unlabeled graph learning? Existing graph neural networks all require labeled nodes from two or more classes
to learn a model.
\end{itemize}

In order to address the above challenges, we propose a novel long-short distance aggregation network (LSDAN) for positive unlabeled (PU) graph learning. 
For \textit{Challenge 1}, we first generate multiple graphs in different hops based on the adjacency matrix, then develop a long-short distance attention model for these graphs.
The long-short distance attention model employs a short-distance attention mechanism to capture the importance of each neighbor node to a target node, and utilizes a long-distance attention approach to model the weights of the different graph with different neighbor nodes for representation learning.
For \textit{Challenge 2}, we employ two \textit{novel risk estimators} for positive unlabeled learning and the expected loss is back-propagated for model learning. 
Experimental results on three real datasets validate the design and effectiveness of our approach.
Our contributions can be summarized below:
\begin{itemize}
\item We first study \textit{positive unlabeled graph learning} for network node classification task (\textit{i.e.} network transductive learning), and present a new deep learning model LSDAN as a solution.
\item We propose a novel attention network for graph data, which captures node significance in both short-distance and long-distance graphs, to model the long-short distance neighboring information in a single layer.
\item Experiments on benchmark graph datasets demonstrate that our graph neural network approach outperforms the baseline methods.
\end{itemize}

The remainder of the paper is structured as follows. Section 2 reviews the related work. Section 3 provides the problem statement. Section 4 presents the proposed algorithm, long-short distance aggregation networks, for PU graph learning. Section 5 illustrates the experimental study, and we conclude this paper in Section 7.

\section{Related work}
This work is closely related to graph neural networks, positive unlabeled learning and PU learning for graph data, which are briefly reviewed below.

\subsection{Graph Neural Networks}
Network node representation aims to map nodes with higher proximities in a network closer to each other in the low-dimensional latent space, which is based on network topology structure only or with side information. For topology structure only embedding methods, most of existing works focused on preserving network structures and properties in embedding vectors \cite{perozzi2014deepwalk} \cite{tang2015line} \cite{kdd_GroverL16}. LINE \cite{tang2015line} and SDNE \cite{Wang_SDNKDD2016} seek to preserve the first-order and second-order proximities between nodes based on the first-order and second-order neighbors. DeepWalk~\cite{perozzi2014deepwalk} employs the random walk sampling strategy to generate the neighborhood of each node. Then, some deep learning  approaches~\cite{cao2016deep,shen2017deep} have been employed to learn more similar feature representations for nodes which can more easily reach each other within $K$ steps. Aside from topology structure only methods, many approaches are proposed to incorporate side information such as node features \cite{pan2018adversarially} \cite{xu2018exploring} \cite{zhang2018anrl}.

Recently, graph neural networks, which are designed to use deep learning architectures on graph-structured data, have drawn significant attention from the research community. Many solutions are proposed to generalize well-established neural network models that work on regular grid structure to deal with graphs with arbitrary structures \cite{wu2019comprehensive,ijcai2019-chun,pan2019learning}.
\citet{Bruna2014Spectral} generalized the convolution operation in the Fourier domain by computing the eigendecomposition of the graph Laplacian.
 Then, a parameterization of the spectral filters with smooth coefficients was proposed to make them spatially localized~\citep{Henaff2015Deep}.  
 \citet{DuvenaudMABHAA15} also considered in the form of spectral analysis, and these networks allowed end-to-end learning of prediction pipelines whose inputs were graphs of arbitrary size and shape. 
 \citet{Defferrard2017Convolutional} proposed to approximate the filters by means of a Chebyshev expansion of the graph Laplacian.
 Finally, \citet{Kipf2016Semi} simplified the previous method by restricting the filters to operate in a $1$-hop neighborhood around each node, which can render the extension of CNN to irregular graphs to learn local and stationary features on graphs. 
 \citet{LiTBZ15} studied feature learning techniques for graph-structured inputs, they modified Graph Neural Networks~\citep{ScarselliGTHM09} to use gated recurrent units and modern optimization techniques and then extended to output sequences. 
 Recently, \citet{HamiltonYL17}  introduced GraphSAGE, a general inductive framework that leverages node feature information to efficiently generate node embeddings for previously unseen data.
 \citet{LiWZH18} proposed a generalized and flexible graph CNN taking data of arbitrary graph structure as input. In that way, a task-driven adaptive graph was learned for each graph data while training.
 \citet{YouLYPL18} proposed Graph Convolutional Policy Network (GCPN), a general graph convolutional network based model for goal-directed graph generation through reinforcement learning. 
 With the widespread application of attention mechanisms, the development of graph attention network methods in graphs has also been promoted. 
 \citet{velickovic2017graph} introduced the attention mechanism to graph neural network through specifying different weights to different nodes in a neighborhood.
 \citet{ZhangSXMKY18} proposed  Gated Attention Networks (GaAN), for learning on graphs. Unlike the traditional multi-head attention mechanism, which equally consumed all attention heads, GaAN used a convolutional sub-network to control each attention head’s importance. %
 Graph neural networks has also been used for cross domain text classification \cite{wu2019domain} or purely unsupervised cross network node classification \cite{wu2020unsupervised}. 
{\color{black}Xu et al. \cite{xu2018powerful} study the expressiveness of graph neural networks in terms of their ability to distinguish any two graphs and introduce Graph Isomorphism Network, which is proved to be as powerful as the Weisfeiler-Lehman test for graph isomorphism. 
You et al. \cite{you2020design} release GraphGym, a powerful platform for exploring different GNN designs and tasks. 
Chen et al. \cite{chen2020iterative} propose an end-to-end graph learning framework, namely Iterative Deep Graph Learning (IDGL), for jointly and iteratively learning graph structure and graph embedding.}

In a recent graph U-Nets design, a gPool~\cite{gao2019graph} procedure is proposed to select top-$k$ nodes to form an induced sub-graph for the next input layer. Although their up-pooling process is efficient, gPool might lose the completeness of the graph structure information, because it only selects top-$k$ nodes, and result in isolated sub-graphs, which hampers the message passing process in subsequent layers. 
In order to model long-distance relationships, GTNs~\cite{yun2019graph} consider all possible meta-paths within a length limit. Instead of using pooling approaches, we develop a long-short distance attention mechanism to model these graphs by generating multiple graphs at different distances based on the adjacency matrix. The long-short distance attention mechanism serves similar purposes as GTNs~\cite{yun2019graph}, but without compromise the graph completeness. 

All existing graph neural networks require users to label data from two or more classes to facilitate the classification task.
To solve this issue, this work aims to  propose a novel long-short distance aggregation network (LSDAN) for positive unlabeled (PU) learning from graphs.

\subsection{Positive Unlabeled Learning}
Positive unlabeled (PU) learning learns a binary classifier model from  positive ($P$) and unlabelled ($U$) data.  
%
Existing PU methods can be divided into two categories based on how unlabeled data $U$ data are handled.
The first category is referred as the two-step strategy,
which first identifies possible negative ($N$) data in $U$, and then performs the ordinary
supervised (PN) learning from both positive and reliable negative examples~\citep{Li2003Learning}.
The second category is referred to as a direct learning method, and regards $U$ data as $N$ data with smaller weights, which learns classification models from the $P$ and $U$ data directly such as One-class SVM~\citep{Sch2014Estimating}, Biased-SVM~\citep{Liu2003Building}.
However, the former heavily relies on the heuristics in identifying $N$ data, and
the latter heavily relies on different choices of the weights of $U$ data, which is computationally expensive to tune.

%
To deal with this issue, some unbiased PU learning methods~\citep{nipsPlessisNS14,Niu2015Convex,nipsKiryoNPS17} are proposed. The main solution is to adopt some novel risk estimators to avoid the bias for PU classification. Specifically,
{In~\citet{Niu2015Convex}’s work}
, an unbiased risk estimator  is proposed to avoid the intrinsic bias for unbiased PU learning.
Recently, a non-negative risk estimator~\citep{nipsKiryoNPS17} is proposed for PU learning, and it is more robust against overfitting when getting minimized, and thus some flexible models can be used given a limited number of $P$ (positive) data.
%
%
%
%
%
%
%
The methods in \cite{nipsKiryoNPS17, Niu2015Convex} employ different estimators for positive and unlabeled data, but they mainly focus on the non-graph data or non-relational data. 
Different from the earlier papers, we focus on the positive unlabeled problem on graph data, by taking both node features and relationships between them into consideration.
Moreover, although the methods in \cite{nipsKiryoNPS17, Niu2015Convex} tried to tackle the positive unlabeled learning problem, 
they are limited to feature extraction but cannot be employed for graph feature learning. 
In our paper, we mainly focus on the graph data and model the long-short distance neighboring information for each node to obtain node features. Through a novel attention network, our method captures node significance in both short-distance and long-distance graphs, which further enhance the feature learning and positive unlabeled graph learning.
%

\subsection{PU Learning for Graph Data}
A handful of works have studied PU Learning for graph data, but under different problem settings. \citet{Zhao2012Positive} proposed an integrated approach
to  select discriminative features for graph classification based upon  positive and unlabeled graphs.
\citet{wu2017positive} proposed a learning framework for classifying a bag of multiple graphs. They assume each object is represented as a bag of graphs and only partial of bags are positively labeled. So their task is to predict the  class label for a whole graph or a bag of graphs.

{\color{black}Our problem setting and learning framework are fundamentally different from these works in three aspects: 1) Existing PU learning for graph data deal with a graph dataset consisting of many graphs, and the task is to predict the  class label for a whole graph or a bag of graphs (\textit{i.e.} inductive graph learning). Our goal is to deal with a single large graph and classify  nodes in the given graph (\textit{i.e.} transductive graph learning); 2) Existing PU learning on graph are all \textit{shallow} and \textit{biased} models, our algorithm, in comparison, is an \textit{unbiased} and \textit{deep} neural network model; and 3) In order to achieve PU learning for graph node classification, we propose to combine graph feature learning into the  classification task, using specifically designed objective function.}

%
%

\begin{table}[ht]
\begin{small}
\centering
  \caption{Summary of Notations and Symbols.}
\begin{tabular}{c|p{0.6\columnwidth}}
\toprule
Notations  &  Descriptions             \\ 
\midrule
$G=\left (V,E,X, Y \right)$   & An attributed graph \\ 
$V, E$    & Node set and edge set of $G$    \\ 
$P, U$    & Labeled and unlabled node set of $G$    \\ 
$n, |V|=n$  & Number of nodes in $G$ \\
$A \in \mathbb{R}^{n\times n}, A^k\in \mathbb{R}^{n\times n}$ & Adjacency matrix ($A$) and length-$k$ walk matrix ($A^k$) of $G$  \\ 
$X\in \mathbb{R}^{n\times m}$ &  Feature matrix of $G$\\
$\textbf{x}_i \in \mathbb{R}^{m}$& Feature vector of node $v_{i}$. Each node has $m$ dimensional features\\
$y_i\in \mathbb{R}^{\{1,u\}}$ & Label of node  $v_i$. A node is either positive (1) or unlabeled (u). \\
$Y\in \mathbb{R}^{n\times 2}$  & Label matrix of $G$ \\
$\kappa$  & the maximum hops for long-short distance aggregation\\
$\sigma(\cdot)$  &  a non-linear activation function \\
$\textbf{W}^{(1)} \in \mathbb{R}^{d\times m}$ & A shared weight matrix for short-distance feature aggregation learning\\
$\alpha_{i,j}$ & Weight value of the neighbor $v_j$ for node $v_i$\\
$a_{st}(\cdot)$  & Short-distance attention function \\
$\textbf{H}\in\mathbb{R}^{n\times d}$  & Short-distance attention feature embedding results \\
$\textbf{O}\in\mathbb{R}^{n\times m}$  & Long-short distance attention feature embedding results \\
$t_{i,j}$ & Attention coefficient between $v_i$ and $v_j$ \\
$\textbf{T} \in \mathbb{R}^{n\times n}$ & Attention coefficient matrix consisting of $t_{i,j}$ \\
$B^k\in \mathbb{R}^{n\times n}$ & \textcolor{black}{$k-$hop adjacency matrix derived from $A^k$ } \\
$\textbf{W}^{(2)} \in \mathbb{R}^{d\times m}$ & A shared weight matrix for long-short distance feature attention learning\\
$a_{ls}(\cdot)$  & Long-short distance attention function\\
$c_{i}$ & the attention coefficient computed by long-short distance attention function $a_{ls}$ \\
$R(f)$ & the expected loss/risk \\
$R_{p}^{+}(f)$  & the expected loss/risk for positive class\\
$R_{n}^{-}(f)$  & the expected loss/risk for negtive class\\
$\hat{R}_{pn}(f)$ & the empirical loss/risk in traditional binary classification\\
$\hat{R}_{p}^{+}(f)$ & the empirical loss/risk for positive class with loss $\mathcal L(f(o_{i}^{p}),1)$\\
$\hat{R}_{p}^{-}(f)$ & the empirical loss/risk for positive class with loss $\mathcal L(f(o_{i}^{p}),0)$\\
$\hat{R}_{n}^{-}(f)$ & the empirical loss/risk for negtive class\\
$\hat{R}_{pu}(f)$ & the empirical loss/risk for positive and unlabeled learning\\
$\hat{R}_{u}^{-}(f)$ & the empirical loss/risk for unlabelled class\\


\bottomrule
\end{tabular}
\label{tab:notations}
\end{small}
\end{table}

\begin{figure*}[t]
\centering
\includegraphics[scale=0.18]{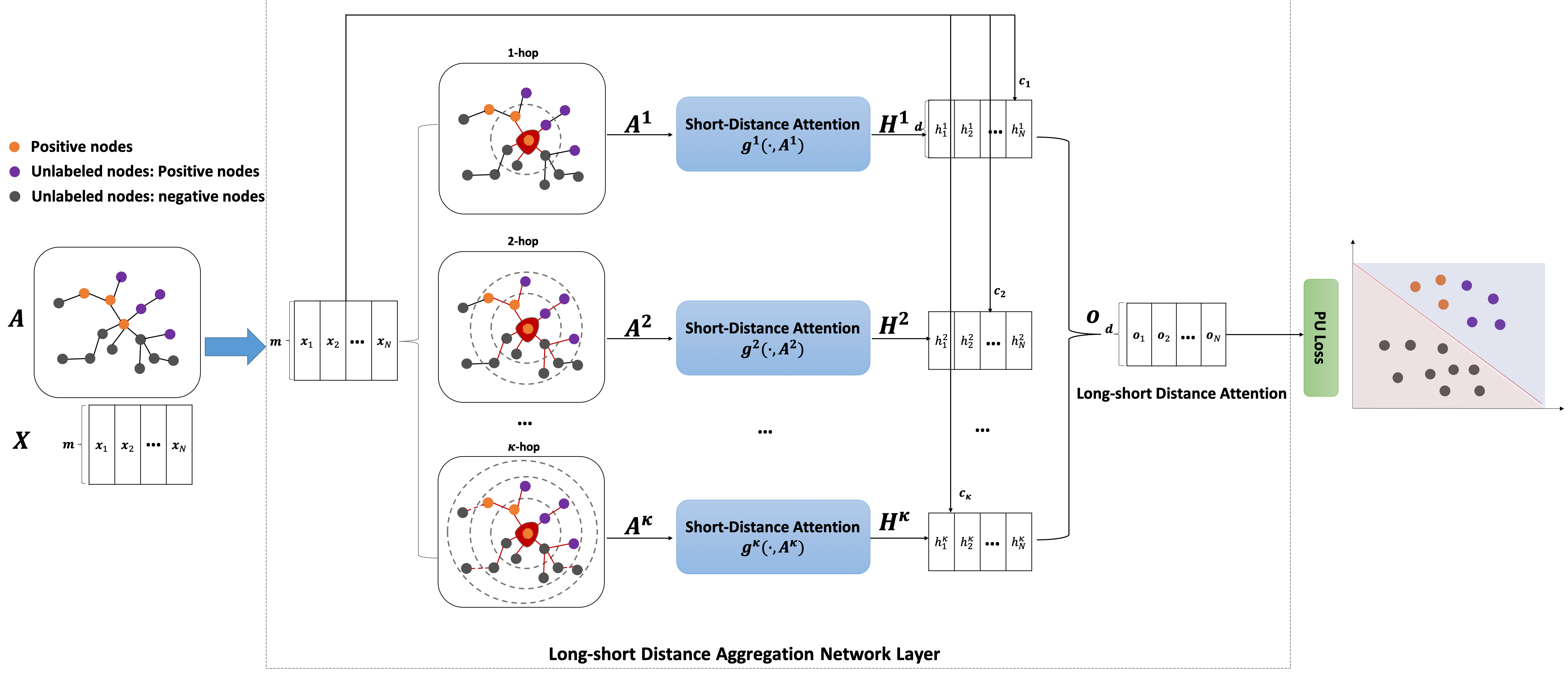}

\caption{The overall architecture of the proposed long-short distance aggregation network (LSDAN) model. 
LSDAN uses higher-order network topology structures and node content ($X$) to progressively learn a long-short distance attention model, whose outputs are integrated into a learning objective function to achieve optimized PU graph learning outcomes. Here, LSDAN uses higher order adjacency matrices to capture \textit{long distance relationship} \textit{w.r.t.} a target node. 
}
\label{fig:The overall architecture of proposed model}

\end{figure*}

\section{Problem Statement}
\paragraph{\textbf{Graph: }}A graph is represented as $G=\left ( V,E,X, Y \right )$, where $V=\left \{ v_{i} \right \}_{i=1,\cdots ,N}$ is a vertex set representing nodes in a graph, and $e_{i,j}=(v_{i},v_{j})\in E$  is an edge indicating relationships between nodes. 
The topological structure of graph $G$ can be represented by an adjacency matrix $A$, where $A_{i,j} = 1$ if $(v_{i},v_{j})\in E$; otherwise $A_{i,j} = 0$. $\textbf{x}_i\in X$ indicates content features associated with each node $v_{i}$. $y_i \in Y=\{+1, 0\}$ is the ground-truth class label for each node, where ``1'' denotes positive class, and ``0'' denotes negative class (not positive). If a node $v_i$ is of interest to a user, then $y_i=1$, or $y_i=0$ otherwise. It is worth noting that although the ground-truth label of each node is binary (1 or 0), only a small portion of positive nodes are labeled in PU graph learning, so the labeled set only has positive samples.

\paragraph{\textbf{Positive Unlabeled Graph Learning (PUGL): }}Assume $V=P \bigcup U$, where $P$ are the labeled nodes ($\forall v_i \in P$, $y_i = 1$) and $U$ are unlabeled nodes. Given a graph  $G=\left (V,E,X, Y \right )$, Positive Unlabeled Graph Learning (PUGL) \textbf{aims} to learn a binary classifier model, $f:\left ( A,X; P \right ) \mapsto Y$, to predict the class labels for unlabeled nodes $U$. In this paper, we propose the first deep learning model for PUGL.

For ease of understanding, Table \ref{tab:notations} summarizes major symbols and notations used in the paper.

\section{Long-short Distance Aggregation Networks for PU Graph Learning}
In this section, we present our proposed LSDAN algorithm for PU Graph learning.
Our learning objectives are to (1) capture the \textit{long-short distance relationship} between nodes, and (2) enable PU learning on a graph.
We will  first present our long-short distance attention network which exploits both {short-distance} and {long-distance} attention for \textit{long-short distance relationship} modeling. Then we present two risk estimators for PU learning.
Our framework, as shown in Figure~\ref{fig:The overall architecture of proposed model}, mainly consists of three components:
\begin{itemize}
    \item \textbf{Short-Distance Attention}. For the input $X$ and an adjacent matrix $A$, a short-distance self attention mechanism is applied to learn a representation for each node.
    \item \textbf{Long-short Distance Attention}.
    Given an input graph $G$, we will first generate multi-hop graph representation based on adjacent matrix $A^1, A^2,\cdots, A^\kappa$. The matrix $A^k$ captures the neighbors in the $k$-th hop of the graph $G$.
    We develop a long-distance attention approach to automatically determine the weights of different graphs $A^1, A^2,\cdots, A^\kappa$.
    \item \textbf{Positive Unlabeled Learning}. Based on our long-short distance attention model, we develop a deep architecture for learning the graph representation of each node. Then the unbiased risk estimator and the non-negative risk estimator are used to estimate the classification loss, respectively. The loss is further back-propagated to the learning progress in an end to end learning framework.
\end{itemize}


\subsection{Short-Distance \textit{vs.} Long-Distance}
\begin{definition}{\textsc{Short-Distance:}} Short-distance is defined as the distance from direct (1-hop) neighbor nodes to a target node.
\end{definition}
The (normalized) adjacency matrix A characterizes the first-order proximity to model the direct relationship (1-hop) between vertices.

\begin{definition}{\textsc{Long-Distance:}} Long-distance is defined as the distances of $k$-hop neighbors ($k>1$) to a target node.
\end{definition}

\begin{figure}[h]
\centering
\includegraphics[scale=0.5]{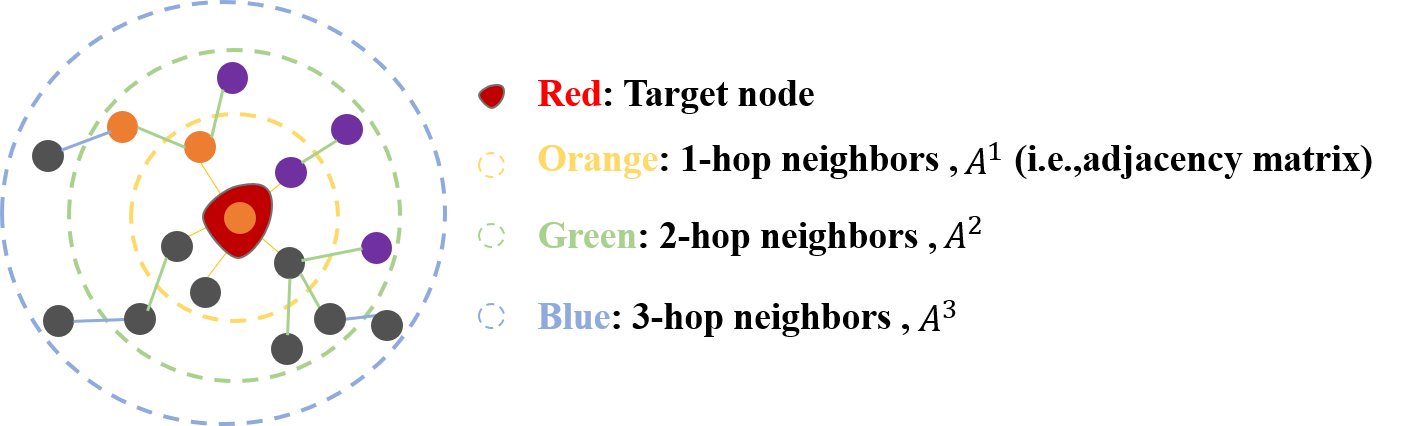}

\caption{\small A conceptual view of long-shot distance attention mechanism. Short-distance corresponds to the 1-hop neighbors which is captured by the adjacency matrix A. Long-distance is defined as the distances from $k$-hop neighbors ($k>1$) to a target node.
}
\label{fig:hops}
\end{figure}

In order to capture long-distance relation for each node, we propose to consider $k$-distance (with varying $k\in[1,\kappa]$)  relational information from the network for graph learning.
Given an input graph $G$, we will first generate multi-hop graph representation based on adjacent matrix $A^1, A^2,\cdots, A^\kappa$. The matrix $A^k$ captures the neighbors in the $k$-th hop of the graph $G$, as shown in Figure~\ref{fig:hops}.
Therefore,
the $k$-distance relationship can be captured by:
\begin{equation}\label{eq:equ1}\small
\setlength{\abovedisplayskip}{0.5mm}
\setlength{\belowdisplayskip}{0.5mm}
A^{k}=\underbrace{A\cdot A\cdots A}_{k},
\end{equation}
where $A_{i,j}^k$ refers to the $k$-hop link relation between node $v_{i}$ and $v_{j}$. In other words, if $A_{i,j}^k\ne 0$, it means that node $v_{i}$ and $v_{j}$ have a $k$-hop relation, or zero otherwise. 

%
%
%

\subsection{Long-short Distance Attention}
In this subsection, we propose to differentiate and combine short-distance attention and long-distance attention to learn network feature representation PU learning. 
\subsubsection{Short-Distance Attention}
Given the input $X\in \mathbb{R}^{n\times m}$ denoting content matrix for all $n$ nodes where each node has $m$ dimension feature, and an adjacent  matrix $A\in \mathbb{R}^{n\times n}$,
a short-distance self attention mechanism is applied to learn a representation for each node, which  aims to capture the node features of the whole graph with short distance by aggregating feature values within each node's neighborhood. 
%
Specifically, 
the input is a set of node features,   $X =\{\bf{x_1},  \bf{x_2},...,\bf{x_N}\}$, $\bf{x_i} \in \mathbb{R}^{m}$ denotes feature vector for node $v_i$ where $m$ denotes the number of input features of each node.
The output of the short-distance attention is a new set of node features,  $\textbf{H}\in\mathbb{R}^{n\times d} =\{\bf{h_1},  \bf{h_2},...,\bf{h_n}\}$, where $\bf{h_i} \in \mathbb{R}^{d}$ and $d$ denotes the number of  embedding features of each node.
%



%
 \begin{equation}\label{eq:k_hop}
 \small
 \setlength{\abovedisplayskip}{1mm}
 \setlength{\belowdisplayskip}{1mm}
 \bf h_{i}= g \left ( \sum_j \alpha_{i,j}A_{i,j}W^{(1)}\textbf{x}_{j}  \right ),
 \end{equation}
%
%
%
where $\bf g$ is a non-linear activation function, $A_{i,j}$ serves as a mask to only aggregate node $V_i$'s direct neighbors (short-distance neighbors) for feature learning. $\alpha_{i,j}$ is weight value capturing the importance of neighbor $v_j$ for node $v_i$.
To automatically learn the parameter $\alpha_{i,j}$, a short-distance self-attention mechanism is developed.
%


To compute $\alpha_{i,j}$, a shared linear transformation is applied to each node through multiply a shared weight matrix $W\in \mathbb{R}^{d\times m}$ in the initial step.
Then an attention coefficient $t_{i,j}$ is computed by an attention function $a_{tt}(\cdot)$: 
\begin{equation}\label{eq:att_a}
\setlength{\abovedisplayskip}{1mm}
\setlength{\belowdisplayskip}{1mm}
t_{i,j}={a{_{tt}}}(W^{(1)}\textbf{x}_{i},W^{(1)}\textbf{x}_{j}),
\end{equation}
which measures the importance of vertex $j$ to vertex $i$.
In the most general formulation, the model allows every node to attend on every other node, dropping all structural information. We inject the graph structure into the mechanism by performing masked attention, masking out all other nodes except direct neighbors based on the adjacency matrix $A$. 
%
%


Furthermore, in order to make coefficients comparable among vertices, a softmax function is utilized to normalize attention coefficients:
\begin{equation}\label{eq:att_alpha}\small
\setlength{\abovedisplayskip}{1mm}
\setlength{\belowdisplayskip}{1mm}
\alpha _{i,j}=\textrm{softmax}_{j}\left ( t_{i,j} \right )=\frac{exp\left ( t_{i,j} \right )}{\sum _{\jmath}{A_{i,\jmath} exp\left (t_{i,\jmath} \right )}},
\end{equation}
In the experiment, the attention mechanism $a_{tt}$ is instantiated with a dot product (parametrized by a weight vector $\textbf{r}\in \mathbb{R}^{2d}$) and a LeakyReLU \citep{xu2015empirical} nonlinearity. 
Fully expanded out, the normalized attention coefficients can be expressed as:
\begin{equation}\label{eq:equ4}\small
\alpha_{i,j}=\frac{{exp}\left (\textrm{LeakyReLU}\left ( \textbf{r}^{T}\left [ W^{(1)} {\bf x_{i}}\oplus W^{(1)} {\bf{x_{j}}} \right ] \right ) \right )}{\sum _{\jmath}{A_{i,\jmath}exp}\left ( \textrm{LeakyReLU}\left ( \textbf{r}^{T}\left [ W^{(1)} {\bf{x_{i}}}\oplus W^{(1)} {\bf{x_{k}}} \right ] \right ) \right )},
\end{equation}
where $a\oplus b$ denotes the concatenation operation of vector $a$ and $b$.

\begin{figure}[h]
\centering
\includegraphics[scale=0.3]{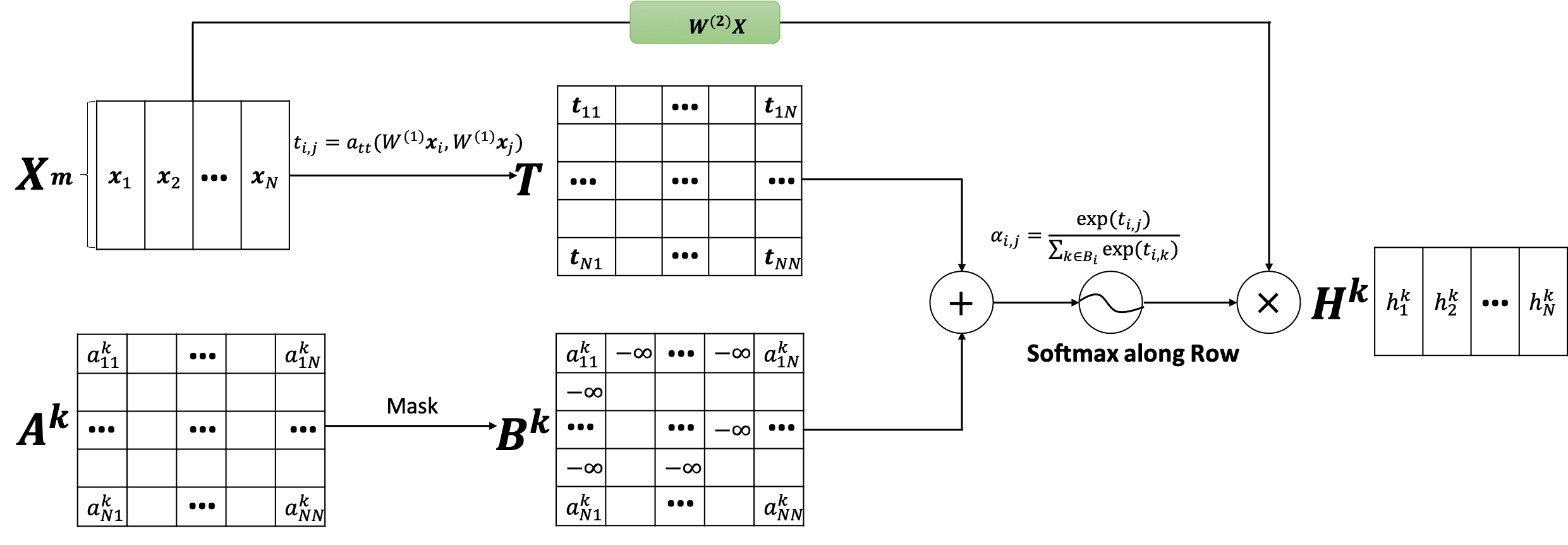}

\caption{\small The mechanism of the long-distance attention, which takes $X$ and $A^k$ as inputs, and computes the importance of nodes via Eq. (\ref{eq:att_a}) to produce the attention coefficient $T$ and get a matrix $B^k$ by masking out all the other nodes except neighbors based on the $A^k$. After that, it will produce a new set of node features $H^k$. 
\label{fig:The Mechanism of the Masked Self Attention}
}

\end{figure}

\subsubsection{Long-Distance Attention}
%
%
%
%
To capture long-distance relation between nodes, we need to aggregate embedding from different range of neighborhoods to produce a unified representation. Because neighbors from different distances contribute differently to learning the representation, we propose a \textit{Long-Distance Attention} scheme to capture the significance of each $k$-hop distance graph neighbors.
%

Specially,  for each  $A^k, k\in\{{1,\cdots,\kappa}\}$, we will perform the self attention to learn the embedding $H^{k}$ for each node (as shown in Figure~\ref{fig:The Mechanism of the Masked Self Attention}). In order to obtain $k$-hop neighborhood for each node, we define a matrix $B^k$ to capture whether two nodes ($v_i$ and $v_j$ are $k$-hop neighbors).

\begin{equation}\label{eq:B}
  B^k_{i,j} =
    \begin{cases}
      1 & \text{if $A^k_{i,j}\ne 0$}\\
      0 & \text{Otherwise}.\\
    \end{cases}       
\end{equation}

By using $B^k$ to mask out other nodes except $k$-hop neighbors, we can learn $k$-hop neighbor aggregated feature embedding as follows. 
\begin{equation}\label{eq:k_hop_k}
\small
\setlength{\abovedisplayskip}{1mm}
\setlength{\belowdisplayskip}{1mm}
\bf{h^k_{i}}=g\left ( \sum_{j} \alpha_{i,j}B^k_{i,j}W^{(1)}\textbf{x}_{j}  \right ),
\end{equation}

We then use the original input $X =\{\bf{x_1},  \bf{x_2},...,\bf{x_n}\}$, $\bf{x_i} \in \mathbb{R}^{m}$ as the key of the attention mechanism, and  perform attention on each graph output 
 $H^{k} =\{\bf{h_1^k},  \bf{h_2^k},...,\bf{h_N^k}\}$, $\bf{h_i^k} \in \mathbb{R}^{d}$,
an attention coefficient $c^{k}_i$ is computed by an attention function $a_{ls}()$: $\mathbb{R}^{d}\times \mathbb{R}^{d}\rightarrow \mathbb{R}$:
\begin{equation}\label{eq:equ7}
\setlength{\abovedisplayskip}{1mm}
\setlength{\belowdisplayskip}{1mm}
c^{k}_i={a_{ls}}({\bf h^{k}_i}, W^{(2)}{\bf x_i}),
\end{equation}
where $W^{(2)}$ is a shared weight matrix for long-short feature attention learning, characterizing the consistency between short-distance and long-distance aggregated features ($W^{(2)}$ also enforces the input ${\bf x^{k}_i}$ of node $i$ to have the same dimension as the $k-$hop embedding features ${\bf h^{k}_i}$). In this paper, we denote $a_{ls}()$ as a dot-product attention function. After that, we further normalize the weight $c^k_i$ with a softmax layer.
\begin{equation}\label{eq:att_ck}\small
\setlength{\abovedisplayskip}{1mm}
\setlength{\belowdisplayskip}{1mm}
c ^{k}_i=\frac{exp\left ( c^{k}_i \right )}{\sum _{k=1}^Kexp\left ( c^{k}_i \right )}.
\end{equation}

After implementing the attention, final embedding output $\mathbf{O}\in \mathbb{R}^{n\times d} = \left  \{ {\bf o_{1},\cdots ,o_{n}} \right \}, {\bf o_{i}}\in \mathbb{R}^{d}$:
\begin{equation}\label{eq:equ8}
\small
\setlength{\abovedisplayskip}{1mm}
\setlength{\belowdisplayskip}{1mm}
{\bf o_{i}} =\sum_{k=1}^{\kappa} {c^{k}_i}{\bf{h^{k}_i}}.
\end{equation}

\subsection{Deep Long-short Distance Aggregation Networks}
\begin{figure*}
\centering
\includegraphics[scale=0.22,width=14cm,height=2cm]{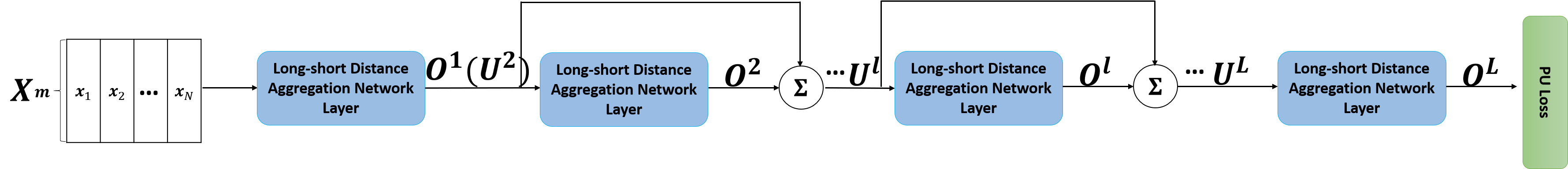}
  \caption{The architecture of multiple layers of the proposed model.}
\label{fig:The architecture of Multiple Layers of the Proposed Model}
\end{figure*}

The short-distance attention and long-distance attention components are integrated into a unified layer, \textit{ Long-short Distance Aggregation Network Layer} (LSDAN), which serves as a building block to construct a deep architecture for node classification in a single network, 
as shown in  Figure \ref{fig:The architecture of Multiple Layers of the Proposed Model}.
%
The LSDAN layers are stacked in the following way:
\begin{itemize}

\item [$\bullet$]
The input $U^{l+1}$ to the $(l+1)$th layer  is the sum of the output $O^{l}$ and the input $U^{l}$ from layer $l$:
\begin{equation}\label{eq:equ9}
\setlength{\abovedisplayskip}{1mm}
\setlength{\belowdisplayskip}{1mm}
U^{l+1}=U^{l}+O^{l}.
\end{equation}

\item [$\bullet$]
A \textit{residual connection} \citep{he2016deep} around two sub-layers and the multi-graph information $A^1, A^2,\cdots, A^\kappa$ are used in different layers.
The residual connection method  provides the input without any transformation to the output of the $(l+1)$th layer, which makes the  $(l+1)$th layer learn something new about the network.
%

\item [$\bullet$]
At the first layer of the network ($l$ = 1),
let $U^{1}=X$,
we will not use the residual connection (i.e. $U^{2}=O^{1}$), because 
we need to first map a high-dimensional node representation to a low-dimensional representation.

\item [$\bullet$]
At the last layer of the network ($l$ = L),
we do not use the residual connection, and map the embedding of nodes to the 2-dimensional representation for PU classification.

\end{itemize}

%

%
%
By this way, we can build arbitrary deep long-short distance aggregation networks to effectively learn graph representation, by leveraging the long-short distance neighboring information.

\subsection{Positive and Unlabeled Graph Learning}
After integrating the structure and content information via a deep long-short distance aggregation network, we will  obtain the new representation $O^L=  \{ {\bf o}_{1}^L,\cdots , {\bf o}_{N}^L  \}, {\bf o}_{i}^L\in \mathbb{R}^{2}$ in the final layer. One key question has arisen as \textit{how can we perform positive unlabelled learning from this new representation}? 

We first formulate the traditional binary classification problem as a risk minimization problem, and then employ two effective positive unlabeled learning methods to approximate the risk for PUGL. 
%
\textit{An unbiased risk estimator} and \textit{a non-negative risk estimator} are used to approximate the risk for PU learning, respectively. By minimizing the risk,
our model is learned in an end-to-end manner.


\vspace{1mm}\noindent
\subsubsection{Traditional Binary Classification}
Given a set of the obtained  representations $O^L=  \{ {\bf o}_{1}^L,\cdots ,{\bf o}_{N}^L  \}$, ${\bf o}_{i}^L\in \mathbb{R}^{2}$ where ${\bf o}_{\text{i}}$ is new feature representation of node $i$. 
In the traditional binary classification,
we need to learn a model $f: \mathcal{O}\rightarrow\mathcal{Y}$, to classify each node ${\bf o}_{\text{i}}$ into the predefined categories $\mathcal{Y} = \{+1, 0\}$, which is the ground-truth label of the node ($1$ denotes positive samples,  $0$ denotes negative samples).

Let  $\mathcal L: \mathbb R \times  \{+1,0\} \rightarrow \mathbb R$ be a loss function, then $\mathcal L(y',y)$ measures the predicting loss for an output $y'$ when the ground truth is $y$.   Let $f$ be a mapping function, and $f(o)$ maps the input $o$ in the range (0,1).
The traditional binary classification problem is formulated as a risk minimization problem:
\begin{equation}
\small
\setlength{\abovedisplayskip}{1mm}
\setlength{\belowdisplayskip}{1mm}
R(f)=\mathbb{E}\left [ \mathcal L(f(O),Y) \right ]=\pi_{p}{R}_{p}^{+}(f)+\pi_{n}{R}_{n}^{-}(f),
\end{equation}
where  $R_{p}^{+}(f) =\mathbb{E}_{p}\left [ \mathcal L(f(O),+1) \right ]$ and $R_{n}^{-}(f) =\mathbb{E}_{p}\left [  \mathcal L(f(O),0) \right ]$ are the expected loss for positive and negative samples. Here, we denote $\pi _{p}=p(Y=+1)$ be the \textit{class-prior probability}, $\pi _{n}=p(Y=0) = 1-\pi _{p}$. $\pi _{p}$ is assumed to known throughout the paper, and it can be estimated from positive data \citep{jain2016estimating}.

Therefore, for traditional binary classification problem i.e., positive and negative learning (PN Learning), we can  minimize an approximated $R(f)$ by, 
\begin{equation}\label{eq:equ10}\small
\setlength{\abovedisplayskip}{1mm}
\setlength{\belowdisplayskip}{1mm}
\hat{R}_{pn}(f)=\pi_{p}\hat{R}_{p}^{+}(f)+\pi_{n}\hat{R}_{n}^{-}(f),
\end{equation}
where  $\hat{R}_{p}^{+}(f)=(1/n_{p})\sum_{i=1}^{n_{p}}\mathcal L(f({\bf o}_{i}^{p}),+1)$ and $\hat{R}_{n}^{-}(f)=(1/n_{n})\sum_{i=1}^{n_{n}}\mathcal L(f(o_{i}^{n}),0)$. Here, $n_{p}$ and $n_{n}$ denote the number of positive/neagtive samples, respectively.

\subsubsection{Unbiased Risk Estimator for PU Learning}
For positive unlabeled (PU) learning, however, negative training data is unavailable. As a result, we need to estimate $\hat{R}_{n}^{-}(f)$ via Eq. (\ref{eq:equ10}).  
Here, we use a novel unbiased risk estimator to approximate  $\hat{R}_{n}^{-}(f)$~\citep{nipsPlessisNS14}.
Specifically,  the expected loss $\hat{R}_{n}^{-}(f)$ of negative samples can be obtain by the expected loss $\hat{R}_{p}^{-}(f)$ of positive samples and the expected loss $\hat{R}_{u}^{-}(f)$ of unlabeled samples, and  is defined as
\begin{equation}\label{eq:equ111}\small
\setlength{\abovedisplayskip}{1mm}
\setlength{\belowdisplayskip}{1mm}
\pi_{n}\hat{R}_{n}^{-}(f) = -\pi_{p}\hat{R}_{p}^{-}(f)+\hat{R}_{u}^{-}(f),
\end{equation}
where $\hat{R}_{p}^{-}(f)=(1/n_{p})\sum_{i=1}^{n_{p}}\mathcal L(f(o_{i}^{p}),0)$, and $\hat{R}_{u}^{-}(f)=(1/n_{u})\sum_{i=1}^{n_{u}}\mathcal L(f(o_{i}^{u}),0)$. Here, $n_{p}$ and $n_{u}$ denote the number of positive/unlabeled samples, respectively.

Therefore, for positive unlabeled (PU) learning, the risk $R(f)$ can be approximated by, 
%
%
\begin{equation}\label{eq:equ11}\small
\setlength{\abovedisplayskip}{1mm}
\setlength{\belowdisplayskip}{1mm}
\hat{R}_{pu}(f)=\pi_{p}\hat{R}_{p}^{+}(f)-\pi_{p}\hat{R}_{p}^{-}(f)+\hat{R}_{u}^{-}(f).
\end{equation}

\vspace{2mm}
\subsubsection{Non-negative Risk Estimator for PU Learning}
Although the unbiased risk estimator can efficiently solve the positive unlabeled learning problem. However,
Eq. (\ref{eq:equ11})  may cause the risk value get negative, because there's a negative sign in front of $\hat{R}_{p}^{-}(f)$. The $\hat{R}_{p}^{-}(f)$ refers to the sample in the positive sample set which is predicted to get the expected risk value with the negative label through the model, that is: if the prediction is the negative sample, loss is 0; if the prediction is not the negative sample, loss is positive. 
This will cause overfitting problem for PU Learning. Motivated by \citet{kiryo2017positive}, we employ a  non-negative risk estimator $\hat{R}_{pu}(f)$, given as follows, 
\begin{equation}\label{eq:equ12}\small
\setlength{\abovedisplayskip}{1mm}
\setlength{\belowdisplayskip}{1mm}
\hat{R}_{pu}(f)=\pi_{p}\hat{R}_{p}^{+}(f)+\max\left \{ 0 , \hat{R}_{u}^{-}(f)-\pi_{p}\hat{R}_{p}^{-}(f)\right \}.
\end{equation}

In our paper, we will minimize the unbiased empirical risk and non-negative empirical risk, respectively. 
Specifically, let $f$ be a mapping function, and we use a Sigmoid activation function $f(o) = \frac{1}{1+ exp(-o)}$  to map the input $\textbf{o}$, which can be learned by our  graph neural network module in Eq. (\ref{eq:equ8}), to the range (0,1).
The Logistic Loss is used in the paper, and 
the loss function $\mathcal L(y'_i,y_i)$ of each sample is defined as:
\begin{equation}\label{eq:equ20}\small
\mathcal L(y'_i,y_i) = -\left[ y_{\text{i}}* log(y'_{\text{i}}) + (1 - y_{\text{i}}) * log(1 - y'_{\text{i}}) \right],
\end{equation}
where $y'_i$ and $y_i$ are the model predict score and the ground-truth for each sample.
The expected loss/risk can be computed by the unbiased empirical risk and non-negative empirical risk via Eq.~(\ref{eq:equ11}) and 
Eq.~(\ref{eq:equ12}), respectively. 
%

\subsection{Algorithm Description}
Our algorithm is illustrated in  Algorithm 1. 
Given a graph $G=\left (V,E,X, Y \right )$, the goal of Positive Unlabeled Graph Learning (PUGL) is to learn the node representations and learn a binary classifier model, $f:\left ( A,X; P \right ) \mapsto Y$, to predict class labels for unlabeled nodes in $U$. 

The algorithm first obtains adjacency matrix $A$ and its power matrices $A^k$ from $G$ by Eq.~(2) (Step 1).
After that, it uses labelled node set $P$ and unlabelled node set $U$ to  calculate \textit{class-prior probability} $\pi_p$ from $G$  (Steps 2-4).
Next, the algorithm carry out long-short distance aggregation to effectively learn node representations by leveraging the long-short distance neighboring information. The final output of node representations are denoted by $O^L$ (Steps 5-18).
Finally, the algorithm formulates a binary classification task as a risk minimization problem, and employs two effective positive unlabeled learning methods to approximate the risk for PUGL. An unbiased risk estimator and a non-negative risk estimator are used to approximate the risk for PU learning, respectively. By minimizing the risk, the loss can be further back-propagated in our proposed model to guide the representation learning to achieve better PU graph learning results (Steps 19-23).

\begin{algorithm}[tb]
	\caption{\small LSDAN for Positive Unlabeled Graph Learning}
	

	\begin{algorithmic}[1]
    \REQUIRE ~~\\
    	 (1) Graph:  $G=\left (V,E,X, Y \right )$\\
    	 (2) Maximum hops for long-short distance aggregation: $\kappa$\\
    	 (3) Maximum graph neural network layers: $L$
    \ENSURE ~~\\
        A binary classifier model, $f:\left ( A,X; P \right ) \mapsto Y$.
		\STATE $\{A,\cdots,A^\kappa\} \leftarrow $ Obtain adjacency matrix $A$ and its power matrices $A^k$ from $G$
		\STATE $P \leftarrow \{v_i| v_i \in V, y_i = 1$\}. Labelled node set. \\
		\STATE $U \leftarrow \{V\backslash P\}$.~~ Unlabelled node set. \\
		\STATE $\{\pi_p\} \leftarrow $ calculate \textit{class-prior probability} from $G$. \\
		\WHILE{not convergence}
		    \FOR{graph neural network layer $l$=1 to $L$}
                \IF{first layer $l==1$}
                    \STATE  $\textbf{V}^1\leftarrow X$
                \ELSIF{last layer $l==L$}
                    \STATE  $\textbf{V}^L \leftarrow \textbf{O}^{L-1}$
                \ELSE
                    \STATE $\textbf{V}^{l} \leftarrow \textbf{V}^{l-1}+\textbf{O}^{l-1}$ 
                \ENDIF
                \FOR{each hop distance $k$=1 to $\kappa$}
		        \STATE  $\textbf{H}^{l,k} \leftarrow $ Learn embedding for each $A^k$ by Eq.~(\ref{eq:k_hop})
		        \ENDFOR
		       \STATE $\textbf{O}^{l} \leftarrow $ Learn the new output representation based on the $H^{l}$  by Eq.~(\ref{eq:equ8})
		    \ENDFOR
		    \IF{\textit{Unbiased Positive Learning}}
                \STATE $[W^{(1)},W^{(2)}]\leftarrow $ Back-propagate loss gradient using unbiased empirical risk Eq.~(\ref{eq:equ11}) and Eq.~(\ref{eq:equ20})
            \ELSIF{\textit{Non-negative Positive Learning}}
                \STATE $[W^{(1)},W^{(2)}]\leftarrow $ Back-propagate loss gradient using non-negative empirical risk Eq.~(\ref{eq:equ12}) and Eq.~(\ref{eq:equ20})                
            \ENDIF
		\ENDWHILE

	\end{algorithmic}
\end{algorithm}

\subsection{Time Complexity Analysis}
Given a graph $G=\left ( V,E,X, Y \right )$, the proposed Long-short Distance Aggregation Network Layer (LSDAN) consists of two parts: $\kappa$ short-distance attentions, and long-short distance attention, where $\kappa$ denotes the number of hops.
It is worth noting that the calculation of the matrix $A^k$ can be done in advance, and there is no need to recalculate the matrix $A^k$ during the training process. The calculation of $A^k$ requires $O(|V|^3)$ time complexity. 
The time complexity for computing each short-distance attention is $O(|V|md+|E|d)$,  where $|V|$ and $|E|$ are the numbers of nodes and edges in the graph, respectively, and $m$ and $d$ denote the dimensions of the input feature and output feature of a single layer, respectively. 
The time complexity of long-short distance attention is $O(md)$.
Therefore, the overall time complexity of the proposed Long-short Distance Aggregation Network Layer (LSDAN) is $O(\kappa|V|^3 + \kappa|V|md+\kappa|E|d+md)$.

In real-world networks, the number of nodes $|V|$ and the number of edges $|E|$ are much larger than feature dimension $m$, embedding size $d$, and the the maximum hops for long-short distance aggregation $\kappa$, where $V$ and $E$ are more than thousands, and $m$, $d$, and $\kappa$ are in hundreds maximum. In addition, we know that $|E|<|V|^2$. Therefore, LSDAN's complexity is asymptotically bounded by $O(\kappa|V|^3)$.


\section{Experiments}
In this section, we conduct experiments to evaluate our model against state-of-the-art algorithms on three real-world datasets. Furthermore, we also provide detailed experimental analysis to show more insights of our model.

\subsection{Experiment Setting}

\noindent\textbf{Datasets}
We employ three widely used citation network datasets (Cora, Citeseer, DBLP) for node classification \citep{yang2015network,Pan2016Tri}.
The details of the experimental datasets are displayed in Table \ref{tab:datasets}.
The Cora dataset contains 2708 nodes, 5429 edges with 7 classes and 1433 features.
The Citeseer dataset contains 3312 nodes, 4732 edges with 6 classes and 3703 features.
The DBLP dataset contains 5818 nodes, 3633 edges with 4 classes and 1587 features.
%
As these datasets have multiple classes, we 
select the class with the relatively large number of samples as P  (positive) class, and all the other classes are regarded as N (negative) class. 
Specifically, for the Cora, Citeseer, and DBLP datasets, we select the class with the label is 3, 2, and 1, respectively, as P (positive) class and the remaining classes as N (negative) class.
After selecting the positive class, we convert the original classification problems of each dataset into binary classification tasks.

\begin{table}[th]
	\centering
  \caption{Statistics of three datasets.}
  \small
  \begin{tabular}{|l|c|c|c|c|c|c|c|c|}\hline
    Dataset & Node & Edges & Classes  & Features    \\\hline
    Cora   & 2708    & 5429  & 7 & 1433 \\\hline
    Citeseer   & 3312     & 4732  & 6 & 3703  \\\hline
    DBLP   & 5818  & 3633  & 4 & 1587  \\\hline
    \end{tabular}
	\label{tab:datasets}
\end{table}




\noindent\textbf{Baselines}
To the best of our knowledge, there is no existing study on positive unlabeled graph neural network learning. To make a fair comparison and evaluate the effectiveness of our design, we select the following baselines with necessary adaption.

We first compare our model with the classical PU learning methods, which focus on the one-step strategy and two-step strategy.
\begin{itemize}
  \item [$\bullet$] \textbf{OC-SVM:}  OC-SVM~\citep{Sch2014Estimating} (One-class SVM  algorithm) is a classical machine learning algorithm based on support vector machine. It only uses positive examples from the node content to build a binary classifier. In our experiments, we use node features as the input of One-class SVM.

\item [$\bullet$] \textbf{LINE\_OC-SVM:}  We first use an unsupervised network embedding method (LINE \cite{tang2015line}) to learn node representation. After that, the learned features are used as the input of One-class SVM.

  \item [$\bullet$] \textbf{GAE\_OC-SVM:}  We use an unsupervised graph embedding method (Graph Auto-Encoders, GAE \cite{kipf2016variational}) to learn node representation by using both the adjacency matrix of nodes and  the features of nodes. After that, the learned features are used as the input of One-class SVM.
  
    \item [$\bullet$] \textbf{Roc-SVM:}  Roc-SVM~\citep{Li2003Learning} uses two step strategies to build a classifier from the node content, and combines the Rocchio method and the SVM technique for PU learning algorithm. In our experiments, we use the features of nodes as the input of Roc-SVM.
    
    \item [$\bullet$] \textbf{LINE\_Roc-SVM:}  We use an unsupervised network embedding method (LINE \cite{tang2015line}) to learn node representation by only using the adjacency matrix of nodes. After that, the learned features are used as the input of Roc-SVM.
  
  \item [$\bullet$] \textbf{GAE\_Roc-SVM:}  We use an unsupervised graph embedding method (Graph Auto-Encoders, GAE \cite{kipf2016variational}) to learn node representation by using both the adjacency matrix of nodes and  the features of nodes. After that, the learned features are used as the input of Roc-SVM.
\end{itemize}

In addition the  above baseline, we also compare our algorithm with different deep learning models. Note that we have integrated the \textit{unbiased risk estimator} and \textit{non-negative risk estimator} into the following models for PU learning.
\begin{itemize}
  \item [$\bullet$]  \textbf{FC:} Full-connected network only applies node features to a multiple layer perceptron (MLP) to learn node representation without using the adjacency matrix of nodes.
  
    \item [$\bullet$]  \textbf{FS:} Full-connected self-attention network uses the node features with a self-attention network to obtain the representation without the adjacency matrix of nodes.

  \item [$\bullet$] \textbf{GCN:}
  GCN uses the graph convolutional network~\citep{Kipf2016Semi} to integrate structure and content information of nodes to learn node representation using the adjacency matrix of nodes.

  \item [$\bullet$]  \textbf{GAT:}  
  GAT uses the graph attention nets~\citep{velickovic2017graph} to exploit  structure and content information of nodes to obtain node representation using the adjacency matrix of nodes. {\color{black}Note that, in the experiments, we only utilize one attention head.}
  
 \item [$\bullet$] {\color{black}\textbf{GATH:}  
  GATH uses the graph attention nets~\citep{velickovic2017graph} to exploit  structure and content information of nodes to obtain node representation using the adjacency matrix of nodes. In addition, multi-head attention is further utilized to stabilize the learning process and encapsulate detailed information about the neighborhood.}
  
  

\end{itemize}

\noindent\textbf{Our method:}
\begin{itemize}
    \item [$\bullet$] 
    
    \textbf{LSDAN\_UPU:}  
  LSDAN\_UPU employs a long-short distance aggregation network to exploit structure and content information of nodes to obtain the final graph representation, and the unbiased risk estimator is utilized for PU learning.
  
    \item [$\bullet$]  \textbf{LSDAN\_NNPU:}  
  LSDAN\_NNPU employs a long-short distance aggregation network to exploit structure and content information of nodes to obtain the graph  representation, and the non-negative risk estimator is utilized for PU learning.

\end{itemize}



%

\vspace{2mm}
\noindent\textbf{Experimental Setup} For fairness of comparison, we randomly split each PN dataset into positive and unlabeled set. Following ~\citet{kiryo2017positive}, we sample  $N_{PN}$ (the total number of positive nodes) nodes from $N$ as negative class. Then we select $p*{N_{PN}}$ nodes from $P$ as the training set, the rest positive nodes 
and negative nodes are used as the unlabeled set (
$p$ is the percentage of training (positive) nodes).
We conduct $10$ trials of randomly splitting, and report the average \textbf{F1 score} 
as final experimental results. 
%

All models were implemented in TensorFlow with the Adam optimizer with a learning rate of $1e^{-4}$ for $500$ steps. 
For parameter setting,
we set the embedding dimension of nodes to $64$ for all methods. We choose $2$ layers for GCN-PU and GAT-PU, where the first GCN/GAT layer contains $64$ hidden units, and the second layer contains $2$ hidden units for classification.
%
%
For the proposed LSDAN, the number hops $\kappa$ is set to $4$.
{\color{black}The number of heads for the multi-head attention mechanism for GATH is set as 8.}



\subsection{Experimental Results}

\begin{table*}[th]
	\centering
  \caption{F1 scores on the Citeseer network using classical positive unlabeled learning methods. Average F1 score and standard
deviation are reported for $10$ random seeds.}
    \small
\begin{tabular}{|c|c|c|c|c|c|c|}
\hline
\%p & OC-SVM & LINE\_OC-SVM & GAE\_OC-SVM & Roc-SVM &  LINE\_Roc-SVM & GAE\_Roc-SVM \\ \hline
\%0.01 & 0.023$\pm$0.004  & 0.041$\pm$0.011  & 0.517$\pm$0.099  & 0.018$\pm$0.005    & 0.177$\pm$0.036   & 0.393$\pm$0.019   \\ \hline
\%0.02 & 0.038$\pm$0.011  &  0.196$\pm$0.028  & 0.614$\pm$0.028  & 0.057$\pm$0.007   & 0.314$\pm$0.024   & 0.464$\pm$0.015   \\ \hline
\%0.03 & 0.054$\pm$0.015  &  0.304$\pm$0.028  & 0.631$\pm$0.024  & 0.079$\pm$0.007   & 0.381$\pm$0.039   & 0.485$\pm$0.020   \\ \hline
\%0.04 & 0.090$\pm$0.009  &  0.361$\pm$0.024  & 0.646$\pm$0.013  & 0.115$\pm$0.009   & 0.424$\pm$0.026   & 0.493$\pm$0.019   \\ \hline
\%0.05 & 0.089$\pm$0.017  &  0.430$\pm$0.024  & 0.658$\pm$0.011  & 0.146$\pm$0.016   & 0.446$\pm$0.027   & 0.500$\pm$0.017   \\ \hline
    \end{tabular}
	\label{tab:F1 on Citeseer1}
\end{table*}


\begin{table*}[th]
	\centering
  \caption{F1 scores on the Citeseer network using unbiased risk estimator for PU learning. Average F1 score and standard
deviation are reported for $10$ random seeds. The best results are reported in boldface.}
    \small
\begin{tabular}{|c|c|c|c|c|c|c|}
\hline
\%p & FC\_UPU & FS\_UPU & GCN\_UPU & GAT\_UPU & {\color{black}GATH\_UPU} & LSDAN\_UPU \\ \hline
\%0.01 & 0.515$\pm$0.112  & 0.456$\pm$0.141  & 0.337$\pm$0.166   & 0.531$\pm$0.083 & 0.564$\pm$0.086  & {\bf0.647$\pm$0.073}     \\ \hline
\%0.02 & 0.522$\pm$0.109  & 0.553$\pm$0.051  & 0.411$\pm$0.139   & 0.578$\pm$0.067 & 0.606$\pm$0.070  & {\bf0.697$\pm$0.067}     \\ \hline
\%0.03 & 0.533$\pm$0.094  & 0.578$\pm$0.026  & 0.496$\pm$0.084   & 0.582$\pm$0.038 & 0.614$\pm$0.042  & {\bf0.710$\pm$0.054}     \\ \hline
\%0.04 & 0.560$\pm$0.060  & 0.629$\pm$0.017  & 0.540$\pm$0.061   & 0.645$\pm$0.017 & 0.668$\pm$0.024  & {\bf0.717$\pm$0.051}     \\ \hline
\%0.05 & 0.588$\pm$0.042  & 0.681$\pm$0.010  & 0.590$\pm$0.055   & 0.696$\pm$0.019 & 0.711$\pm$0.016  & {\bf0.738$\pm$0.036}     \\ \hline
    \end{tabular}
	\label{tab:F1 on Citeseer2}
\end{table*}

\begin{table*}[th]
	\centering
  \caption{F1 scores on the Citeseer network using non-negative risk estimator for PU learning. Average F1 score and standard
deviation are reported for $10$ random seeds. The best results reported in boldface.}
    \small
\begin{tabular}{|c|c|c|c|c|c|c|}
\hline
\%p      & FC\_NNPU & FS\_NNPU & GCN\_NNPU & GAT\_NNPU & {\color{black}GATH\_NNPU} & LSDAN\_NNPU \\ \hline
\%0.01 & 0.684$\pm$0.013   & 0.682$\pm$0.007   & 0.433$\pm$0.258    & 0.775$\pm$0.030 & 0.777$\pm$0.029    & {\bf0.786$\pm$0.043}      \\ \hline
\%0.02 & 0.626$\pm$0.054   & 0.695$\pm$0.008   & 0.564$\pm$0.300    & 0.775$\pm$0.024 & 0.779$\pm$0.022   & {\bf0.804$\pm$0.028}      \\ \hline
\%0.03 & 0.710$\pm$0.016   & 0.705$\pm$0.009   & 0.623$\pm$0.259    & 0.796$\pm$0.017 & 0.803$\pm$0.015   & {\bf0.813$\pm$0.014}      \\ \hline
\%0.04 & 0.734$\pm$0.013   & 0.725$\pm$0.009   & 0.721$\pm$0.199    & 0.814$\pm$0.015 & 0.815$\pm$0.009   & {\bf0.828$\pm$0.009}      \\ \hline
\%0.05 & 0.743$\pm$0.015   & 0.745$\pm$0.008   & 0.812$\pm$0.011    & 0.830$\pm$0.011 & 0.832$\pm$0.008   & {\bf0.840$\pm$0.007}      \\ \hline
    \end{tabular}
	\label{tab:F1 on Citeseer3}
\end{table*}

\begin{table*}[th]
	\centering
  \caption{F1 scores on the DBLP network using  classical positive unlabeled learning methods. Average F1 score and standard
deviation are reported for $10$ random seeds.}
    \small
    \begin{tabular}{|c|c|c|c|c|c|c|c|}\hline
\%p & OC-SVM & LINE\_OC-SVM & GAE\_OC-SVM & Roc-SVM &  LINE\_Roc-SVM & GAE\_Roc-SVM \\ \hline
\%0.01 & 0.445$\pm$0.029  &  0.349$\pm$0.035  & 0.576$\pm$0.033  & 0.056$\pm$0.012  & 0.355$\pm$0.017   & 0.515$\pm$0.037       \\ \hline
\%0.02 & 0.543$\pm$0.009  &  0.471$\pm$0.023  & 0.624$\pm$0.017  &0.144$\pm$0.026   & 0.463$\pm$0.029   & 0.567$\pm$0.038     \\ \hline
\%0.03 & 0.580$\pm$0.004  &  0.519$\pm$0.018  & 0.637$\pm$0.012  &0.234$\pm$0.019   & 0.499$\pm$0.022   & 0.597$\pm$0.025       \\ \hline
\%0.04 & 0.601$\pm$0.006  &  0.547$\pm$0.012  & 0.643$\pm$0.008  &0.314$\pm$0.037   & 0.524$\pm$0.027   & 0.613$\pm$0.013        \\ \hline
\%0.05 & 0.611$\pm$0.006  &  0.567$\pm$0.015  & 0.643$\pm$0.007  &0.371$\pm$0.037   & 0.545$\pm$0.023   & 0.616$\pm$0.017      \\ \hline
    \end{tabular}
	\label{tab:F1 on DBLP1}
\end{table*}

\begin{table*}[th]
	\centering
  \caption{F1 scores on the DBLP network using unbiased risk estimator for PU learning. Average F1 score and standard
deviation are reported for $10$ random seeds. The best results are reported in boldface.}
    \small
    \begin{tabular}{|c|c|c|c|c|c|c|c|c|c|}\hline
\%p & FC\_UPU & FS\_UPU & GCN\_UPU & GAT\_UPU & {\color{black}GATH\_UPU} & LSDAN\_UPU \\ \hline
\%0.01 & 0.510$\pm$0.108  & 0.523$\pm$0.060  & 0.398$\pm$0.134   & 0.547$\pm$0.033 & 0.582$\pm$0.050  & {\bf0.687$\pm$0.056}     \\ \hline
\%0.02 & 0.526$\pm$0.075  & 0.612$\pm$0.022  & 0.501$\pm$0.096   & 0.655$\pm$0.035 & 0.690$\pm$0.031  & {\bf0.709$\pm$0.067}     \\ \hline
\%0.03 & 0.572$\pm$0.042  & 0.645$\pm$0.012  & 0.573$\pm$0.072   & 0.729$\pm$0.023 & 0.734$\pm$0.018  & {\bf0.740$\pm$0.049}     \\ \hline
\%0.04 & 0.610$\pm$0.029  & 0.671$\pm$0.010  & 0.633$\pm$0.054   & 0.749$\pm$0.021 & 0.753$\pm$0.015  & {\bf0.757$\pm$0.035}     \\ \hline
\%0.05 & 0.646$\pm$0.020  & 0.693$\pm$0.010  & 0.669$\pm$0.052   & 0.771$\pm$0.013 & 0.771$\pm$0.012  & {\bf0.777$\pm$0.033}     \\ \hline
    \end{tabular}
	\label{tab:F1 on DBLP2}
\end{table*}

\begin{table*}[th]
	\centering
  \caption{F1 scores on the DBLP network using non-negative risk estimator for PU learning. Average F1 score and standard
deviation are reported for $10$ random seeds. The best results are reported in boldface.}
    \small
    \begin{tabular}{|c|c|c|c|c|c|c|c|c|}\hline
        \%p      & FC\_NNPU & FS\_NNPU & GCN\_NNPU & GAT\_NNPU & {\color{black}GATH\_NNPU} & LSDAN\_NNPU \\ \hline
       \%0.01 & 0.650$\pm$0.032    & 0.677$\pm$0.007  & 0.419$\pm$0.128  & 0.767$\pm$0.019 & 0.775$\pm$0.018 & {\bf 0.808$\pm$0.012}                  \\\hline
       \%0.02 & 0.521$\pm$0.092    & 0.695$\pm$0.023  & 0.599$\pm$0.050  & 0.807$\pm$0.017 & 0.808$\pm$0.014  & {\bf 0.833$\pm$0.015}                  \\\hline
       \%0.03 & 0.710$\pm$0.011    & 0.715$\pm$0.007  & 0.685$\pm$0.032  & 0.824$\pm$0.008 & {\bf 0.825$\pm$0.009}  & 0.824$\pm$0.008               \\\hline
       \%0.04 & 0.597$\pm$0.046    & 0.725$\pm$0.011  & 0.734$\pm$0.026  & 0.836$\pm$0.009 & 0.838$\pm$0.009 & {\bf 0.849$\pm$0.010 }                  \\\hline
       \%0.05 & 0.741$\pm$0.009    & 0.746$\pm$0.009  & 0.760$\pm$0.024  & 0.845$\pm$0.009 & 0.845$\pm$0.008 & {\bf 0.857$\pm$0.010}                  \\\hline
    \end{tabular}
	\label{tab:F1 on DBLP3}
\end{table*}

\begin{table*}[th]
	\centering
  \caption{F1 scores on the Cora network using classical positive unlabeled learning methods. Average F1 score and standard
deviation are reported for $10$ random seeds.}
    \small
    \begin{tabular}{|c|c|c|c|c|c|c|c|}\hline
\%p & OC-SVM & LINE\_OC-SVM & GAE\_OC-SVM & Roc-SVM &  LINE\_Roc-SVM & GAE\_Roc-SVM \\ \hline
\%0.01 & 0.111$\pm$0.213  & 0.101$\pm$0.016  & 0.618$\pm$0.046  & 0.039$\pm$0.008   & 0.229$\pm$0.042   & 0.447$\pm$0.040    \\ \hline
\%0.02 & 0.263$\pm$0.004  & 0.260$\pm$0.031  & 0.698$\pm$0.027  & 0.073$\pm$0.010   & 0.348$\pm$0.032   & 0.497$\pm$0.029      \\ \hline
\%0.03 & 0.293$\pm$0.010  & 0.367$\pm$0.025  & 0.700$\pm$0.019  & 0.128$\pm$0.013   & 0.411$\pm$0.029   & 0.504$\pm$0.015     \\ \hline
\%0.04 & 0.324$\pm$0.014  & 0.433$\pm$0.032  & 0.701$\pm$0.018  & 0.169$\pm$0.017   & 0.443$\pm$0.030   & 0.502$\pm$0.017      \\ \hline
\%0.05 & 0.358$\pm$0.015  & 0.482$\pm$0.026  & 0.708$\pm$0.014  & 0.218$\pm$0.018   & 0.465$\pm$0.017   & 0.516$\pm$0.022     \\ \hline
    \end{tabular}
	\label{tab:F1 on Cora1}
\end{table*}

\begin{table*}[th]
	\centering
  \caption{F1 scores on the Cora network with the unbiased risk estimator for PU learning. Average F1 score and standard
deviation are reported for $10$ random seeds. The best results are reported in boldface.}
    \small
    \begin{tabular}{|c|c|c|c|c|c|c|c|c|c|}\hline
\%p & FC\_UPU & FS\_UPU & GCN\_UPU & GAT\_UPU & {\color{black}GATH\_UPU} & LSDAN\_UPU \\ \hline
\%0.01 & 0.524$\pm$0.098  & 0.503$\pm$0.115  & 0.454$\pm$0.079   & 0.563$\pm$0.080 & 0.597$\pm$0.086  & {\bf0.746$\pm$0.073}     \\ \hline
\%0.02 & 0.531$\pm$0.091  & 0.567$\pm$0.056  & 0.579$\pm$0.059   & 0.693$\pm$0.055 & 0.720$\pm$0.055  & {\bf0.796$\pm$0.050}     \\ \hline
\%0.03 & 0.561$\pm$0.053  & 0.623$\pm$0.024  & 0.650$\pm$0.035   & 0.771$\pm$0.022 & 0.790$\pm$0.017  & {\bf0.824$\pm$0.017}     \\ \hline
\%0.04 & 0.586$\pm$0.036  & 0.661$\pm$0.019  & 0.686$\pm$0.034   & 0.808$\pm$0.020 & 0.821$\pm$0.016  & {\bf0.836$\pm$0.018}     \\ \hline
\%0.05 & 0.614$\pm$0.026  & 0.686$\pm$0.015  & 0.721$\pm$0.021   & 0.829$\pm$0.016 & 0.838$\pm$0.010  & {\bf0.843$\pm$0.013}     \\ \hline
    \end{tabular}
	\label{tab:F1 on Cora2}
\end{table*}

\begin{table*}[th]
	\centering
  \caption{F1 scores on the Cora network with the non-negative risk estimator for PU learning. Average F1 score and standard
deviation are reported for $10$ random seeds. The best results are highlighted in boldface.}
    \small
    \begin{tabular}{|c|c|c|c|c|c|c|c|c|}\hline
        \%p      & FC\_NNPU & FS\_NNPU & GCN\_NNPU & GAT\_NNPU & {\color{black}GATH\_NNPU} & LSDAN\_NNPU \\ \hline
       \%0.01 & 0.542$\pm$0.086    & 0.673$\pm$0.008  & 0.610$\pm$0.248  & 0.772$\pm$0.026 & 0.782$\pm$0.013  &  {\bf0.825$\pm$0.016}          \\\hline
       \%0.02 & 0.593$\pm$0.039    & 0.683$\pm$0.008  & 0.771$\pm$0.078  & 0.817$\pm$0.024 & 0.826$\pm$0.022 &  {\bf0.841$\pm$0.020}          \\\hline
       \%0.03 & 0.641$\pm$0.033    & 0.697$\pm$0.010  & 0.819$\pm$0.041  & 0.842$\pm$0.016 & 0.848$\pm$0.016 &  {\bf0.850$\pm$0.013}          \\\hline
       \%0.04 & 0.666$\pm$0.028    & 0.713$\pm$0.010  & 0.842$\pm$0.018  & 0.859$\pm$0.015 & 0.859$\pm$0.010 & {\bf0.860$\pm$0.013}            \\\hline
       \%0.05 & 0.691$\pm$0.025    & 0.725$\pm$0.008  & 0.850$\pm$0.012  & 0.866$\pm$0.008 & 0.866$\pm$0.007 & {\bf0.867$\pm$0.009}            \\\hline
    \end{tabular}
	\label{tab:F1 on Cora3}
\end{table*}

%

The results of our evaluation experiments are presented in Table \ref{tab:F1 on Citeseer1}, \ref{tab:F1 on Citeseer2}, and Table \ref{tab:F1 on Citeseer3}, Table \ref{tab:F1 on DBLP1}, Table \ref{tab:F1 on DBLP2}, and Table \ref{tab:F1 on DBLP3}, Table \ref{tab:F1 on Cora1}, Table \ref{tab:F1 on Cora2}, and Table \ref{tab:F1 on Cora3}.
From these results, we have the following observations:
 \begin{itemize}
    \item [(1)] In most cases, OC-SVM and Roc-SVM are inferior to other methods. This is because the traditional shallow learning methods do not capture the underlying graph structure information. Besides, we can also find that on the DBLP data set (in Table \ref{tab:F1 on DBLP1}, \ref{tab:F1 on DBLP2}, and  \ref{tab:F1 on DBLP3}), when the value of \%p is small, the performance of OC-SVM may be higher than that of GCN\_NNPU. This may be because it is difficult for the model to learn good representation features for all nodes when there are rather few positive samples.

    \item [(2)] 
    Both GAE\_OC-SVM and GAE\_Roc-SVM outperform OC-SVM and LINE\_OC-SVM, Roc-SVM and LINE\_Roc-SVM, respectively. This shows the effectiveness of unsupervised graph embedding by using both the adjacency matrix of nodes and the features of node. 
    
   \item [(3)]  GAT\_UPU and GAT\_NNPU outperform FC\_UPU, FS\_UPU, and FC\_NNPU, FS\_NNPU, respectively. This confirms that it is useful to take node relationships into consideration for node representation learning.
   {\color{black}Furthermore,  GATH\_UPU and GATH\_NNPU  outperform  GAT\_UPU and GAT\_NNPU, which shows the effectiveness of introducing the multi-head attention.}
   
   \item [(4)] The proposed LSDAN\_UPU and LSDAN\_NNPU outperform GAT\_UPU and GAT\_NNPU which only capture 
   short-distance neighboring information. The results show the effectiveness of our algorithm in exploiting multi-hop neighbors to capture \textit{long-short distance relationship} in graph learning. Meanwhile, our model shows superior performance in positive unlabeled learning problem.

%
    \item [(5)] The results also show that the proposed LSDAN\_UPU and LSDAN\_NNPU consistently outperform all  the other baselines on all three datasets with different training ratios. It demonstrates that  long-short distance aggregation network together with the unbiased and non-negative risk estimators (UPU and NNPU) can better capture data distribution and the underlying relationship among data by integrating the feature information and graph information into a unified framework. 

 \end{itemize}

\subsection{Analysis of Different Components}
As our proposed model contains two key components: the long-short distance aggregation network (LSDAN) and the positive unlabeled (PU) learning component. In this section, we compare variants of the proposed model with respect to the following  aspects to demonstrate the effectiveness of the long-short distance aggregation network and the positive unlabeled (PU) learning component.

The following LSDAN variants are designed for comparison.
\begin{itemize}
    \item LSDAN${\neg p}$: A variant of LSDAN with the positive unlabeled (PU) loss being removed, and only using the cross entropy loss.
    \item LSDAN\_UPU${\neg l}$: A variant of LSDAN\_UPU with the long-short distance aggregation network being removed, and only using the short-distance aggregation layer.
    \item LSDAN\_NNPU${\neg l}$: A variant of LSDAN\_NNPU with the long-short distance aggregation network being removed, and only using the short-distance aggregation layer.
\end{itemize}
The ablation study results are shown in Table \ref{tab:varientscomparisions1}, Table \ref{tab:varientscomparisions2} and Table \ref{tab:varientscomparisions3}.

\begin{table*}[th]
\centering
  \caption{F1 score comparisions between LSDAN variants on the Citeseer network.}
    \small
\begin{tabular}{|c|c|c|c|c|c|}
\hline
\%p    & LSDAN${\neg p}$ & LSDAN\_UPU${\neg l}$ &  LSDAN\_UPU & LSDAN\_NNPU${\neg l}$ & LSDAN\_NNPU \\ \hline
\%0.01 & 0.362 & 0.531 & {\bf0.647} & 0.775 & {\bf0.786} \\ \hline
\%0.02 & 0.373 & 0.578 & {\bf0.697} & 0.775 & {\bf0.804} \\ \hline
\%0.03 & 0.385 & 0.582 & {\bf0.710} & 0.796 & {\bf0.813} \\ \hline
\%0.04 & 0.396 & 0.645 & {\bf0.717} & 0.814 & {\bf0.828} \\ \hline
\%0.05 & 0.409 & 0.696 & {\bf0.738} & 0.830 & {\bf0.840} \\ \hline
\end{tabular}
\label{tab:varientscomparisions1}
\end{table*}

\begin{table*}[th]
\centering
  \caption{F1 score comparisions between LSDAN variants on the DBLP network.}
    \small
\begin{tabular}{|c|c|c|c|c|c|}
\hline
\%p    & LSDAN${\neg p}$ & LSDAN\_UPU${\neg l}$ &  LSDAN\_UPU & LSDAN\_NNPU${\neg l}$ & LSDAN\_NNPU \\ \hline
\%0.01 & 0.431 & 0.547 & {\bf0.687} & 0.767 & {\bf0.808} \\ \hline
\%0.02 & 0.437 & 0.655 & {\bf0.709} & 0.807 & {\bf0.833} \\ \hline
\%0.03 & 0.441 & 0.729 & {\bf0.740 }& 0.824 & {\bf0.824} \\ \hline
\%0.04 & 0.461 & 0.749 & {\bf0.757 }& 0.836 & {\bf0.849} \\ \hline
\%0.05 & 0.472 & 0.771 & {\bf0.777 }& 0.845 & {\bf0.857} \\ \hline
\end{tabular}
\label{tab:varientscomparisions2}
\end{table*}

\begin{table*}[th]
\centering
  \caption{F1 score comparisions between LSDAN variants on the Cora network.}
    \small
\begin{tabular}{|c|c|c|c|c|c|}
\hline
\%p    & LSDAN${\neg p}$ & LSDAN\_UPU${\neg l}$ &  LSDAN\_UPU & LSDAN\_NNPU${\neg l}$ & LSDAN\_NNPU \\ \hline
\%0.01 & 0.337 & 0.563 & {\bf0.746} & 0.772 & {\bf0.825} \\ \hline
\%0.02 & 0.349 & 0.693 & {\bf0.796} & 0.817 & {\bf0.841} \\ \hline
\%0.03 & 0.372 & 0.771 & {\bf0.824} & 0.842 & {\bf0.850} \\ \hline
\%0.04 & 0.394 & 0.808 & {\bf0.836} & 0.859 & {\bf0.860} \\ \hline
\%0.05 & 0.415 & 0.829 & {\bf0.843} & 0.866 & {\bf0.867} \\ \hline
\end{tabular}
\label{tab:varientscomparisions3}
\end{table*}

\begin{figure}[htpb]
\begin{minipage}[htbp]{1\linewidth}
\centering
\subfigure[Citeseer network]{\includegraphics[scale=0.5]{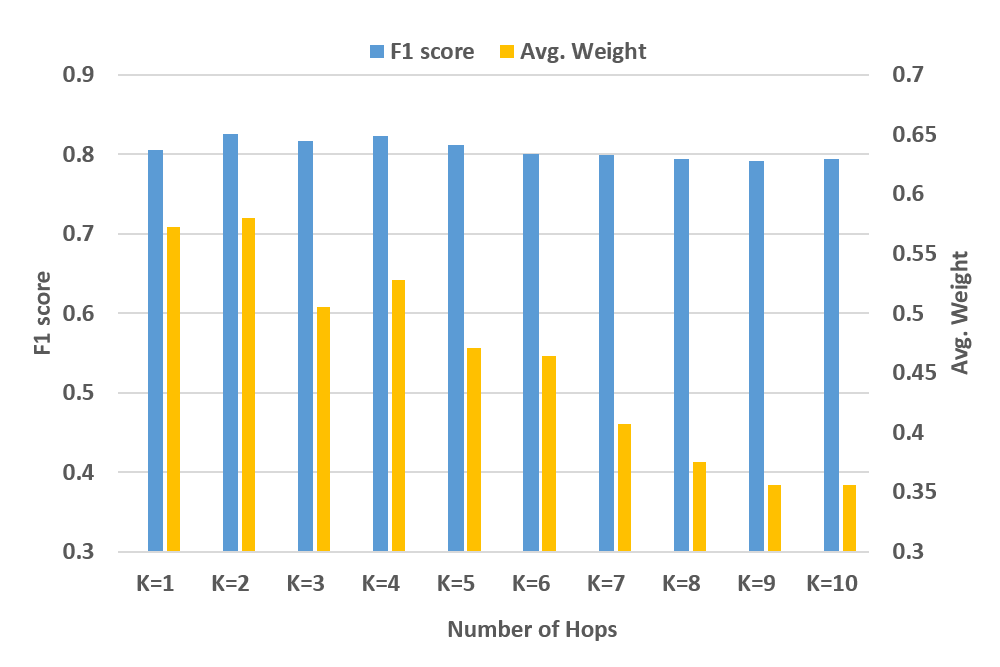}}\
\subfigure[Cora network]{\includegraphics[scale=0.5]{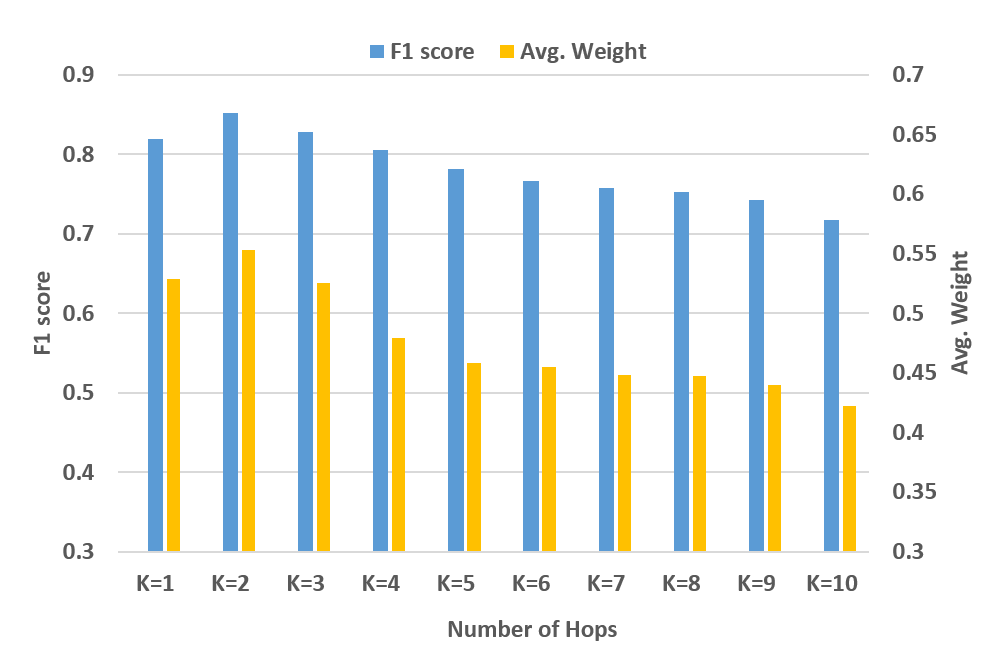}}\
\subfigure[DBLP network]{\includegraphics[scale=0.5]{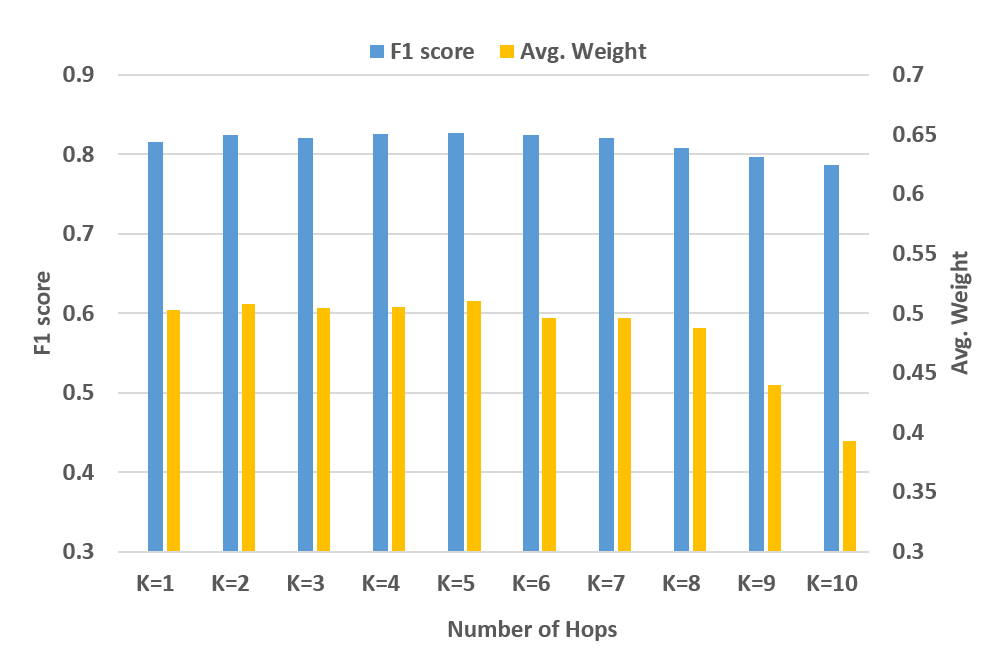}}
\caption{F1 scores and attention weight values by using a single $A^k$ only ($\%p = 0.02$ and $L=2$).}
\label{fig:atten}
\end{minipage}
\end{figure}

\begin{figure}[htpb]
\centering
\subfigure[Embedding dimension $d$]{\includegraphics[scale=0.5]{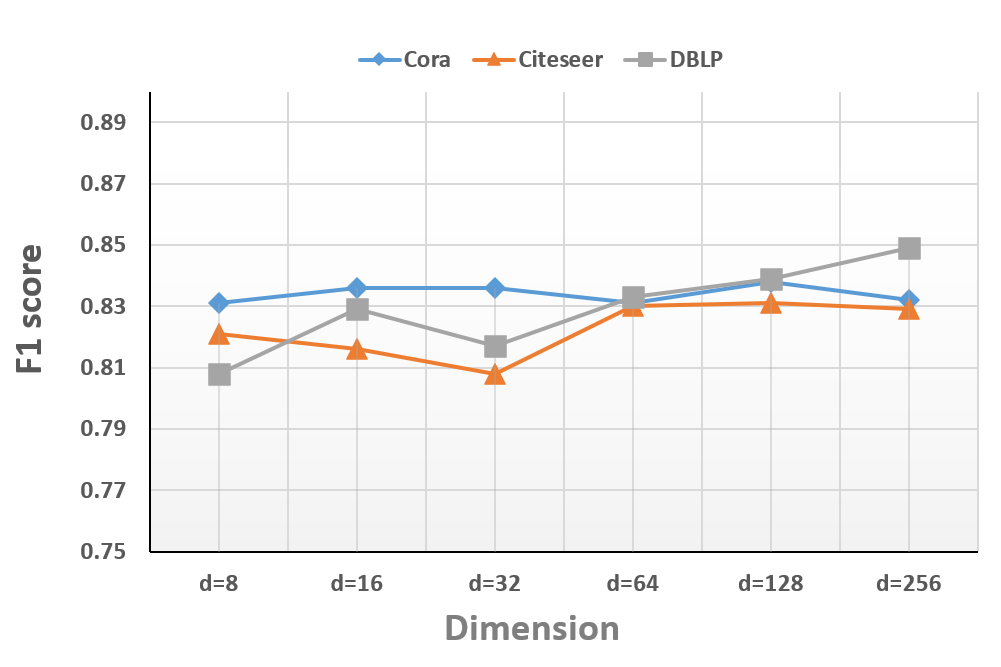}}
\subfigure[Distances at $k$-hops]{\includegraphics[scale=0.5]{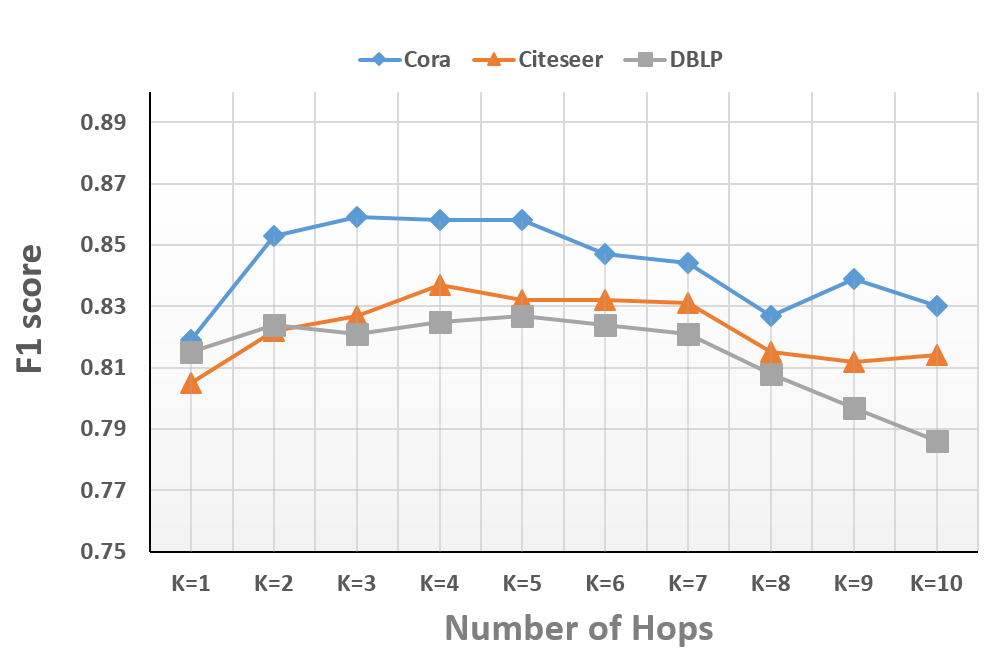}}
\subfigure[Layers $L$]{\includegraphics[scale=0.5]{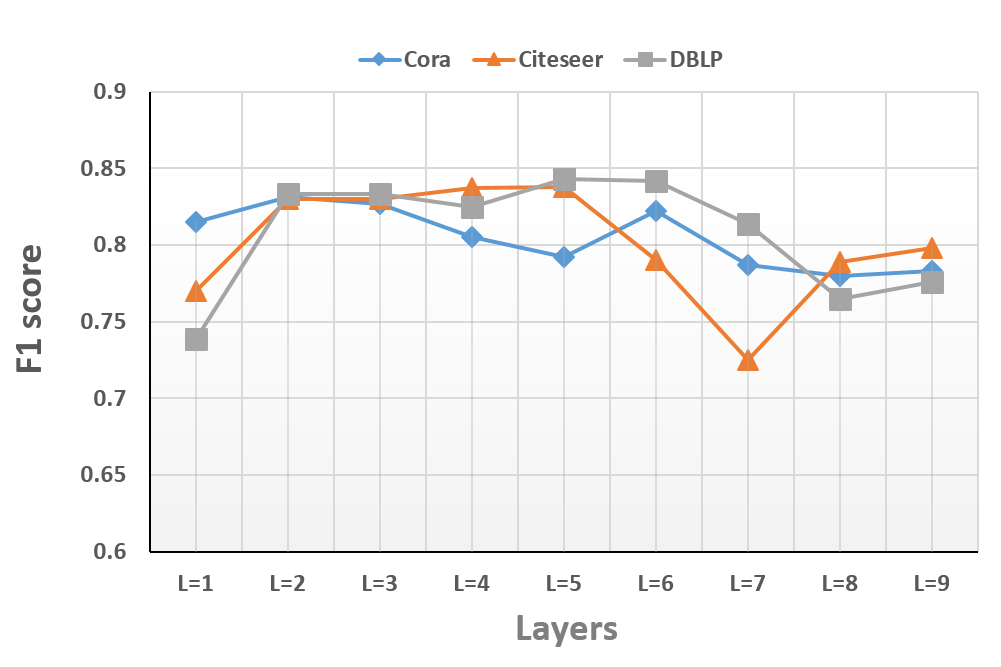}}
\caption{Parameter analysis with respect to embedding dimension $d$, distance at $k$-hops, and layers $L$.}
\label{fig:PA}
\end{figure}

\subsubsection{Impact of the PU loss}
We compare LSDAN${\neg p}$ with LSDAN\_UPU and LSDAN\_NNPU to investigate the effectiveness of the positive unlabeled risk estimators employed in our paper. 
From the result, we find that both LSDAN\_UPU and LSDAN\_NNPU perform better than LSDAN${\neg p}$, which confirms the superiority of the risk estimators for positive unlabeled learning problem.

\subsubsection{Impact of long-short distance aggregation network}
For fair comparisons, in the actual operation, we only employ one attention head for GAT\_UPU and GAT\_NNPU. In other words, LSDAN\_UPU${\neg l}$ are the same as GAT\_UPU, and LSDAN\_NNPU${\neg l}$ are the same as GAT\_NNPU. Therefor, the results of LSDAN\_UPU${\neg l}$ are the the same as that of GAT\_UPU, and same for LSDAN\_NNPU${\neg l}$ \textit{vs.} GAT\_NNPU. In order to verify the effectiveness of the long-short distance aggregation network, we compare LSDAN\_UPU with LSDAN\_UPU${\neg l}$, and LSDAN\_NNPU with LSDAN\_NNPU${\neg l}$.
From Table \ref{tab:varientscomparisions1}, Table \ref{tab:varientscomparisions2} and Table \ref{tab:varientscomparisions3}, we can easily observe that LSDAN\_UPU and LSDAN\_NNPU perform significantly better than LSDAN\_UPU${\neg l}$ and LSDAN\_NNPU${\neg l}$. This confirms that the usage of long-short distance aggregation network can learn a superior representation for nodes in graphs.

\subsection{Analysis of the Learned Long-Distance Attention}
%
%
%
We further  study the learned attentions 
to understand why our \textit{long-distance attention mechanism} helps improve the performance. Specifically, we examine  which graph ($A^k$) attracts more attentions for the classification task.
We report F1 score using $A^k$ only and its attention weight value learned by our LSDAN algorithm.
%
%
For simplicity, we only report the results on LSDAN\_NNPU which are presented in Figure~\ref{fig:atten}.
%
%

The results show that the performance of the single graph ($A^k$) and the long-distance attention value 
positively correlate. For instance, on the Citeseer dataset, $A^2$ receives more weight as its predictive power is stronger.
The results validate that our approach allows different nodes to focus on the different hops based on the adjacency matrix, to achieve better performance.

\subsection{Parameter Analysis}
\textbf{Embedding Dimensions $d$:} We vary $d$ with $\%p=0.02$, $L=2$ and report the results on the three datasets in Fig.~\ref{fig:PA}(a).
We can find that F1 scores show a clear increase from $8$ to $64$ on the Cora and DBLP,
while it decreases slightly in the $32$nd dimension in the Citeseer.
When the number of embedding dimensions continuously increases, the performance starts to remain stable.
This is intuitive as more embedding dimensions can encode more useful information from data.

\textbf{Distance at $\kappa$-Hops :}
We also report F1 scores over different choices of $\kappa$ with $\%p = 0.02$ and  $L=2$ on the three datasets in Fig.~\ref{fig:PA}(b).
It can be seen from Fig.~\ref{fig:PA}(b) that when $k$=4, satisfactory results have been achieved on different data sets. This confirms that the long distance relation is really important to better capture graph structure information, and multiple graphs can learn complementary local information. Simultaneously, when $k$ is greater than 4, as $k$ gets bigger and bigger, the result becomes worse, possible because some redundant information interfere the learning, and the learned $k$-hop relational information becomes less informative for node representation learning.
%

\textbf{Number of layers $L$:}
Fig.~\ref{fig:PA}(c) shows the influence of the number of layers on performance on three datasets.
Here, we set $\kappa=4$, and $\%p = 0.02$.
For the datasets considered here, best results are obtained with a $2$- or $3$-layer model.
We can see that  the setting $L$ = 2 has a significant improvement over the setting $L$ = 1 on three datasets.
We observe that the performance will slightly decrease for models deeper than $6$ layers, this may overfit as the number of parameters increases with model depth.

\section{Discussion}
{\color{black}PU learning is traditionally applied to data with independent and identical distributions (\textit{i.e.} non-relational data). Some earlier works have extended it to graph (relational data) databases. For example, \cite{Zhao2012Positive} proposed an integrated approach
to  select discriminative features for graph classification based upon  positive and unlabeled graphs.
\cite{wu2017positive} proposed a learning framework for classifying a bag of multiple graphs. Therefore, there are existing work which apply PU learning to graphs/networks, but under different learning settings. \\

In our problem settings, although nodes are not independent, we can assume that labels are independently provided for selected nodes (which will be marked as labeled nodes). In other words, when setting a set of nodes to be labeled, a random approach is used to select a small porting of nodes being labeled. In fact, all existing works in node classification (transductive graph learning) employ this setting. Once the labels are provided, graph neural networks here are employed to learn the new representation for each node. After each node is represented into a vector space, this task is similar to general PU learning, and then the risk estimators can be employed for PU learning. \\

One novelty of our approach is that we integrate the feature learning by GNNs and risk estimation into a unified and end-to-end framework. As demonstrated in our algorithm, this approach performs very well in the benchmark datasets.\\

In this paper, we are following common protocols in the research to set up the PU learning for graphs, with a small portion of randomly selected nodes being labeled. However, some nodes may have a higher chance being labeled, due to their linkages or connections. This is, indeed, determined by the applications and network structures. It will be a very interesting direction for future work to investigate further.}


\section{Conclusion}
In this paper, we propose a novel  long-short distance aggregation network (LSDAN) for positive unlabeled graph learning.
We argue that existing algorithms largely overlook the \textit{long-distance relationship}, and only exploit 1-hop neighbors to aggregate information to learn feature representation for nodes.
In order to leverage long-distance relation between nodes, we propose a long-short distance aggregation network to jointly exploit the short-distance and long-short attention from different range of neighborhood to learn feature for each node. 
{\color{black} In addition, two novel risk estimators are proposed for positive unlabeled graph learning.} Experiments and comparisons on three benchmark graph datasets demonstrate the effectiveness of our algorithm.

\begin{acks}
This research is supported by the U.S. National Science Foundation (NSF) through Grant Nos. IIS-1763452, CNS-1828181, and IIS-2027339.
\end{acks}


%
\bibliographystyle{ACM-Reference-Format}
\bibliography{acmart}


\end{document}